\documentclass[10pt,twocolumn,letterpaper]{article}
\usepackage[pagenumbers]{wacv} % To force page numbers, e.g. for an arXiv version
\usepackage{graphicx}
\usepackage{amsmath}
\usepackage{amssymb}
\usepackage{booktabs}
\usepackage{adjustbox}
\usepackage{multirow}
\usepackage{bbding}
\usepackage{pifont}

\usepackage[pagebackref,breaklinks,colorlinks]{hyperref}
\usepackage{orcidlink}
\usepackage[capitalize]{cleveref}
\crefname{section}{Sec.}{Secs.}
\Crefname{section}{Section}{Sections}
\Crefname{table}{Table}{Tables}
\crefname{table}{Table}{Tabs.}

\def\wacvPaperID{1798} % *** Enter the WACV Paper ID here
\def\confName{WACV}
\def\confYear{2025}

\begin{document}

\title{High-Fidelity Document Stain Removal via A Large-Scale Real-World Dataset and A Memory-Augmented Transformer}

\author{Mingxian Li\textsuperscript{1}\footnotemark[1], Hao Sun\textsuperscript{1}\footnotemark[1], Yingtie Lei\orcidlink{0009-0001-8002-2052}\textsuperscript{2}\footnotemark[1], \\Xiaofeng Zhang\orcidlink{0000-0002-7185-4682}\textsuperscript{3}, Yihang Dong\orcidlink{0009-0006-8111-9063}\textsuperscript{4}, Yilin Zhou\textsuperscript{1}, Zimeng Li\orcidlink{0000-0003-2798-3134}\textsuperscript{5}, Xuhang Chen\orcidlink{0000-0001-6000-3914}\textsuperscript{1,2}\footnotemark[2]\\
\textsuperscript{1}School of Computer Science and Engineering, Huizhou University\\
\textsuperscript{2}Faculty of Science and Technology, University of Macau\\
\textsuperscript{3}School of Electronic Information and Electrical Engineering, Shanghai Jiao Tong University\\
\textsuperscript{4}Shenzhen Institute of Advanced Technology, Chinese Academy of Sciences\\
\textsuperscript{5}School of Electronic and Communication Engineering, Shenzhen Polytechnic University\\
{\tt\small \url{https://github.com/CXH-Research/StainRestorer}}
}

\maketitle

\renewcommand{\thefootnote}{\fnsymbol{footnote}}
\footnotetext[1]{Equal contribution.}
\footnotetext[2]{Corresponding author.}

\begin{abstract}
Document images are often degraded by various stains, significantly impacting their readability and hindering downstream applications such as document digitization and analysis. The absence of a comprehensive stained document dataset has limited the effectiveness of existing document enhancement methods in removing stains while preserving fine-grained details. To address this challenge, we construct StainDoc, the first large-scale, high-resolution ($2145\times2245$) dataset specifically designed for document stain removal. StainDoc comprises over 5,000 pairs of stained and clean document images across multiple scenes. This dataset encompasses a diverse range of stain types, severities, and document backgrounds, facilitating robust training and evaluation of document stain removal algorithms. Furthermore, we propose StainRestorer, a Transformer-based document stain removal approach. StainRestorer employs a memory-augmented Transformer architecture that captures hierarchical stain representations at part, instance, and semantic levels via the DocMemory module. The Stain Removal Transformer (SRTransformer) leverages these feature representations through a dual attention mechanism: an enhanced spatial attention with an expanded receptive field, and a channel attention captures channel-wise feature importance. This combination enables precise stain removal while preserving document content integrity. Extensive experiments demonstrate StainRestorer's superior performance over state-of-the-art methods on the StainDoc dataset and its variants StainDoc\_Mark and StainDoc\_Seal, establishing a new benchmark for document stain removal. Our work highlights the potential of memory-augmented Transformers for this task and contributes a valuable dataset to advance future research.
\end{abstract}

\section{Introduction}
\label{sec:introduction}
Document contamination by stains significantly impairs readability and visual quality, hindering research and applications such as Optical Character Recognition (OCR)~\cite{mori1992historical,gupta2007ocr}. Eliminating these contaminants enhances document usability and broadens their application scope, particularly in fields like archaeology~\cite{lombardi2020deep}. However, traditional document enhancement methods face considerable challenges when applied to stained documents, often lacking precision in handling fine-grained information crucial for accurate stain identification and removal~\cite{kligler2018document,sauvola2000adaptive,xiong2018degraded,hedjam2014constrained}. These approaches rely on predefined image features or rules and perform poorly when confronted with complex stain residues, especially when stains overlap with text or image edges~\cite{sharma2001show}. Moreover, they often fail to preserve intricate document details, such as fonts, images, and various graphical elements, during the stain removal process~\cite{moghaddam2009low}.

In recent years, deep neural networks have achieved great success in various visual tasks~\cite{zhang2022correction,liu2023coordfill,jiang2020geometry,jiang2021deep,liu2020fine,li2023cee,yang2024adaptive,li2022monocular,huang2024deformmlp,li2022few,10385149}.
Recent advancements in deep learning have also introduced promising solutions to document enhancement~\cite{hradivs2015convolutional,ma2018docunet,zhao2021scene,gangeh2021end}, enabling the learning of complex document features from large datasets and the removal of degradation while preserving document details and integrity. For instance, DocDiff~\cite{yang2023docdiff} effectively manages complex document details through its high-frequency residual refinement module, while Transformer-based models~\cite{feng2021doctr,souibgui2022docentr,zhang2024docres,liu2023explicit2,zhu2024test,li2024cross,zhang1,zhang2,zhang3,zhang4,zhang5,zhang6,yan2024hierarchical,song2024local,yan2024hierarchical2} leverage self-attention mechanisms to achieve notable results in tasks such as geometric and illumination correction. However, these approaches also face limitations, primarily the requirement for substantial high-quality training data and the potential struggle with certain stain types, such as ink and tea stains~\cite{gupta2016synthetic,burie2015icdar2015,cheng2023m6doc}.

To address these shortcomings, we propose a comprehensive approach that includes the construction of a large-scale, high-resolution dataset specifically designed for document stain removal and the development of StainRestorer, a Transformer-based model that effectively removes stains while preserving fine-grained document information. The StainDoc dataset encompasses various document types and stains, ensuring robust model performance across diverse scenarios. StainRestorer introduces a memory module to enhance the model's feature representation capabilities and its ability to handle complex stains through hierarchical feature extraction and semantic memorization mechanisms. The Stain Removal Transformer (SRTransformer), another key component of the StainRestorer architecture, leverages rich, hierarchical stain representations from the DocMemory module to perform precise stain removal. It employs an enhanced spatial attention mechanism to better capture stain patterns across the document, coupled with a channel attention mechanism that focuses on channel-wise relationships to emphasize crucial features for stain identification and removal. This dual-attention approach enables SRTransformer to effectively remove stains while preserving the integrity of the underlying document content.

Our contributions can be summarized as follows:
\begin{enumerate}
\item We introduce StainDoc, the first large-scale, high-resolution ($2145\times2245$) dataset specifically designed for document stain removal. This dataset comprises over 5,000 pairs of stained and clean document images across multiple scenes, encompassing a diverse range of stain types, severities, and document backgrounds, filling a critical gap in the field of document enhancement.
\item We propose the DocMemory module, which captures hierarchical stain representations at part, instance, and semantic levels by employing a series of Memory Units to extract and analyze deep features at various levels of granularity within the document.
\item We propose the Stain Removal Transformer (SRTransformer), which enhances the model's spatial mapping ability for precise stain removal, effectively distinguishing between stain artifacts and genuine document details, leading to high-quality stain removal while preserving document content integrity.
\end{enumerate}

The remainder of this paper is structured as follows:~\cref{sec:related_work} reviews related work and identifies research gaps.~\cref{sec:staindoc_dataset} introduces the StainDoc dataset, detailing its construction and unique features.~\cref{sec:methodology} presents our proposed StainRestorer methodology.~\cref{sec:experiments} evaluates StainRestorer's performance against state-of-the-art methods. Finally,~\cref{sec:conclusion} concludes the paper.
%%%%%--------------------------------------------------------------------------------------------------------------------------------%%%%%
\section{Related Work}
\label{sec:related_work}
\subsection{Document Enhancement}

Recent advancements in document enhancement have addressed various image degradation issues. DeepOtsu~\cite{he2019deepotsu} introduced an iterative deep learning framework for enhancement and binarization. DocDiff~\cite{yang2023docdiff} presented the first diffusion-based framework for deblurring, denoising, and watermark removal. DocProj~\cite{li2019document} focused on correcting geometric and illumination distortions.

GAN-based approaches have shown promise, with GAN-HTR~\cite{jemni2022enhance} integrating text recognition, DE-GAN~\cite{souibgui2020gan} restoring severely degraded images, and UDOC-GAN~\cite{wang2022udoc} addressing illumination correction using unpaired data.

Other notable contributions include GCDRNet~\cite{zhang2023appearance} for improving camera-captured documents, DocNLC~\cite{wang2024docnlc} for handling multiple degradation types, and Text-DIAE~\cite{souibgui2023text} for joint text recognition and image enhancement.

Despite these advancements, the removal of stains in documents remains largely unaddressed. Our work aims to fill this gap by constructing a new dataset for document image enhancement, including stained documents, and proposing a new method for document stain removal.

\subsection{Vision Transformer}

Vision Transformers have revolutionized many image processing tasks~\cite{liu2023explicit,liu2024depth,zheng2024smaformer,liu2024dh,liu2024forgeryttt,chen2,chen5,chen8}. Dosovitskiy \etal pioneered Transformer's application to image recognition~\cite{dosovitskiy2020image}. The Swin Transformer~\cite{liu2021swin} introduced a hierarchical structure and shifted window approach, later adapted for image restoration in SwinIR~\cite{liang2021swinir}.

Uformer~\cite{wang2022uformer} and Restormer~\cite{zamir2022restormer} further advanced Transformer-based image restoration, focusing on efficient processing of high-resolution images.

In document enhancement field, DocTr~\cite{feng2021doctr} and its improved version~\cite{feng2023deep} addressed geometric and lighting distortions. DocEnTr~\cite{souibgui2022docentr} employed self-attention and patch-based processing for document enhancement, while DocRes~\cite{zhang2024docres} introduced dynamic task-specific prompts for unified document restoration.

Given the demonstrated effectiveness of Vision Transformers in various image processing tasks, we have incorporated this architecture into our proposed method for document stain removal.
%%%%%--------------------------------------------------------------------------------------------------------------------------------%%%%%
\section{StainDoc Dataset}
\label{sec:staindoc_dataset}
To address the critical gap in document stain removal research, we introduce StainDoc, the first comprehensive dataset specifically designed for this task.\begin{figure}[ht]
    \begin{minipage}[b]{1.0\linewidth}
        \begin{minipage}[b]{0.32\linewidth}
            \centering
            \centerline{\includegraphics[width=\linewidth]{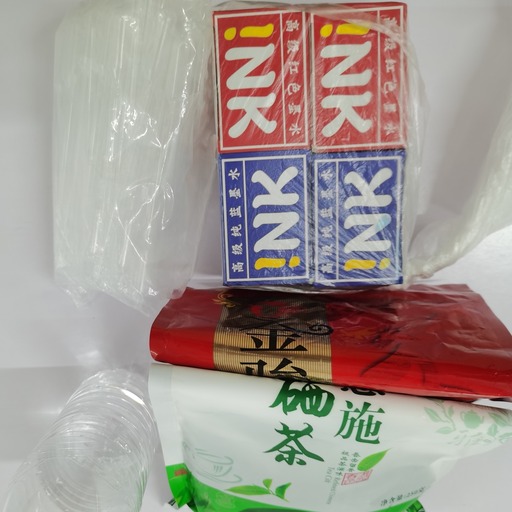}}
            \centerline{(a)}\medskip
        \end{minipage}
        \begin{minipage}[b]{0.32\linewidth}
            \centering
            \centerline{\includegraphics[width=\linewidth]{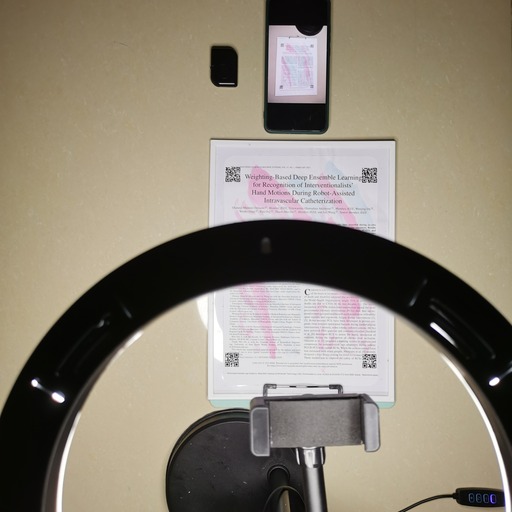}}
            \centerline{(b)}\medskip
        \end{minipage}
        \begin{minipage}[b]{0.32\linewidth}
            \centering
            \centerline{\includegraphics[width=\linewidth]{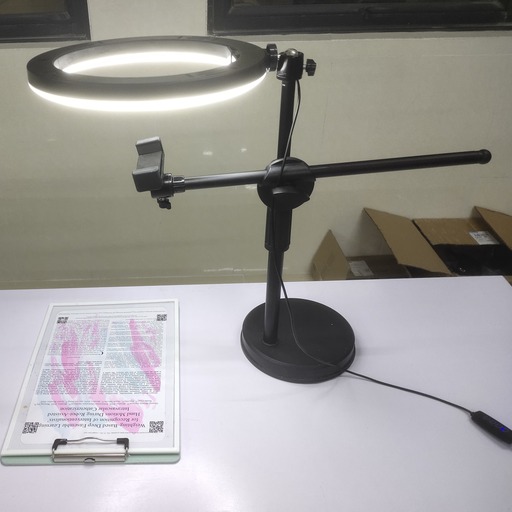}}
            \centerline{(c)}\medskip
        \end{minipage}
    \end{minipage}
    \caption{
    Dataset construction process: (a) Stain application and document preparation, (b) Photography under controlled conditions, (c) Post-processing and standardization.
    }
  \label{fig:eq}
\end{figure}
This dataset comprises over 5,000 pairs of stained and clean document images, enabling the development and evaluation of well-trained models for precise stain removal and restoration of clear, clean original documents.

\cref{fig:eq} illustrates the dataset construction process. We collected around 300 documents with multilingual texts and figures. We simulated common staining factors to replicate real-world scenarios, such as tea, red ink, and blue ink, applying varying degrees of stains to each document. Subsequently, we captured the documents with different levels of degradation under consistent lighting conditions. We then standardized the classification of documents based on stain type and severity. Finally, we screened, cropped, and preprocessed the documents.

To ensure dataset consistency, we utilized standard A4 paper as the document base throughout. We corrected color temperature deviations by setting the white balance, mitigating color casts that occur under specific light sources. Manual focus and ISO value adjustments were employed during photography to maintain consistency in image capture. We selected common staining agents encountered in daily life, specifically red ink, blue ink, black tea, and green tea. These stains can be broadly categorized into two types:

\begin{itemize}
    \item Liquid beverage stains: Tea brewed from green and black tea leaves, characterized by yellow-brown and red-brown colors. These stains contain components like tea polyphenols, causing varying degrees of staining and penetration on the document surface.
    \item Ink stains: Including red and blue ink, which have bright colors and strong permeability, easily forming spots or lines on documents.
\end{itemize}
\begin{figure}[ht]
    \begin{minipage}[b]{1.0\linewidth}
        \begin{minipage}[b]{0.48\linewidth}
            \centering
            \centerline{\includegraphics[width=\linewidth]{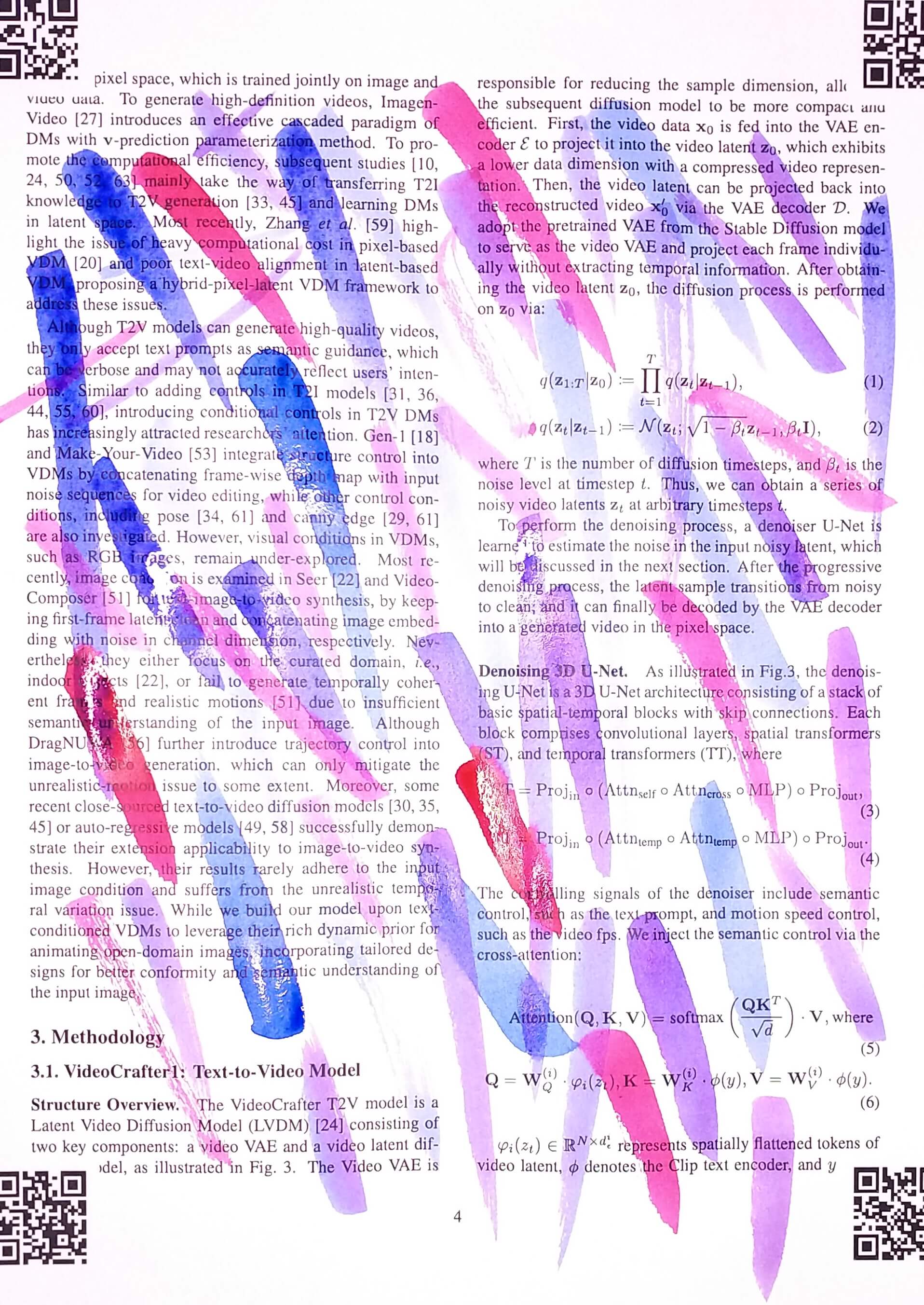}}
            \centerline{(a)}\medskip
        \end{minipage}
        \begin{minipage}[b]{0.48\linewidth}
            \centering
            \centerline{\includegraphics[width=\linewidth]{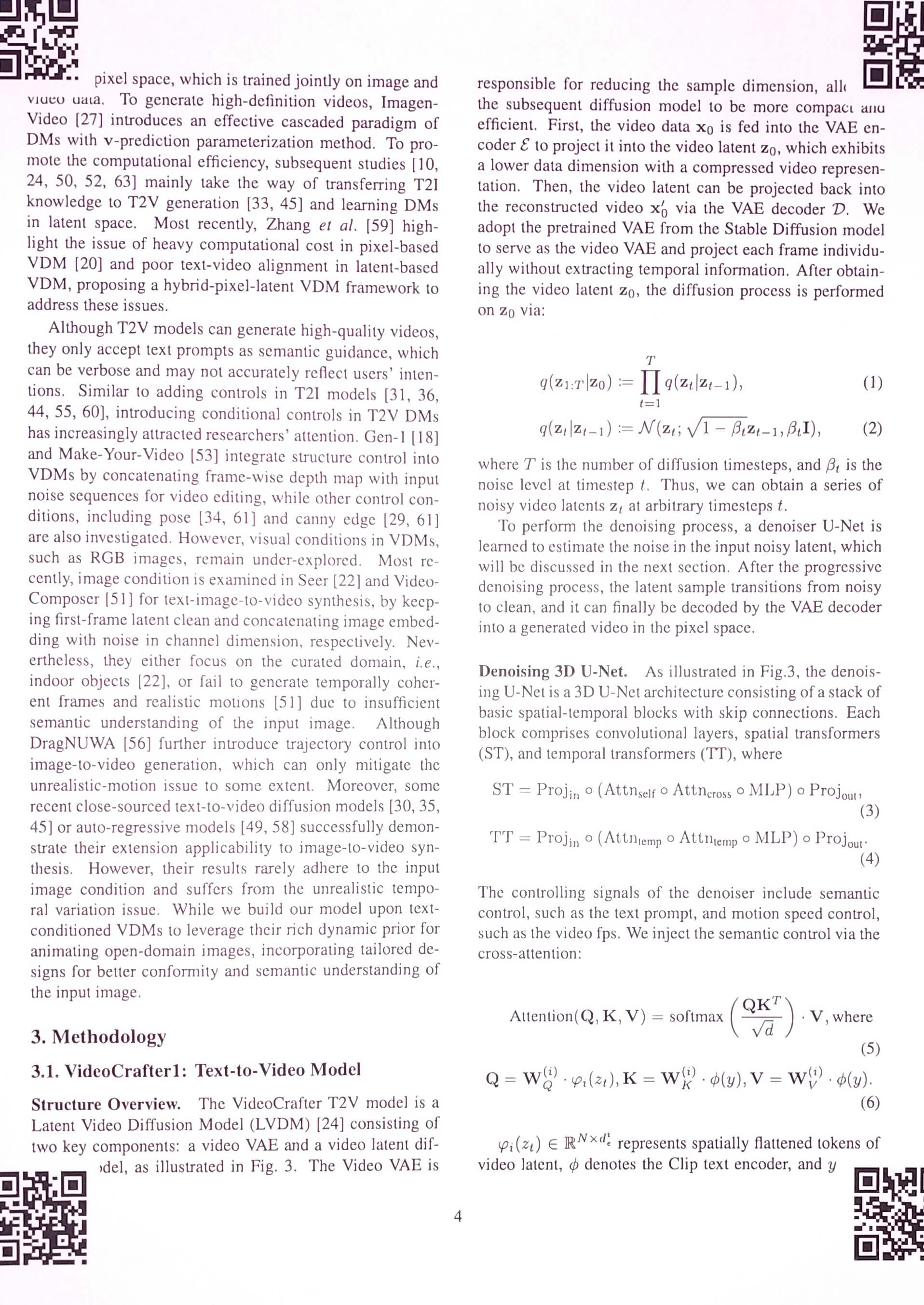}}
            \centerline{(b)}\medskip
        \end{minipage}
    \end{minipage}
    
    \caption{
    A example image pair from the Stain5K dataset: (a) is stained document image and (b) is clean original document image. More dataset samples are available in the supplementary material. 
    }
  \label{fig:sample}
\end{figure}
By selecting these common staining agents, our dataset closely aligns with real-world application scenarios.
Additionally, our dataset encompasses shooting effects from various mobile phone models, including OPPO A52, HUAWEI nova 6, Redmi Note 13Pro+, iPhone 11 Pro Max, and VIVO Z5x. The camera aspect ratio was set to 4:3 to ensure clarity and uniform processing.
In the final stage of dataset formation, we conducted rigorous screening to ensure dataset integrity and quality. We eliminated samples with poor shooting quality (\eg, blurry images, low contrast, and significant detail loss) and filtered out data with extremely high content similarity to maintain diversity. After cropping out-of-boundary areas, we derived a total of 5,060 image pairs with a mean resolution of $2145\times2245$. A example image pair is shown in~\cref{fig:sample}.

%%%%%--------------------------------------------------------------------------------------------------------------------------------%%%%%
\section{Methodology}
\label{sec:methodology}
This section details the architecture and training methodology of StainRestorer. StainRestorer integrates a memory-augmented module (DocMemory) with a Stain Removal Transformer (SRTransformer) to effectively address the challenges of real-world document stain removal.
\begin{figure*}[t]
    \begin{minipage}[b]{1.0\linewidth}
        \centering
            \centerline{\includegraphics[width=\linewidth]{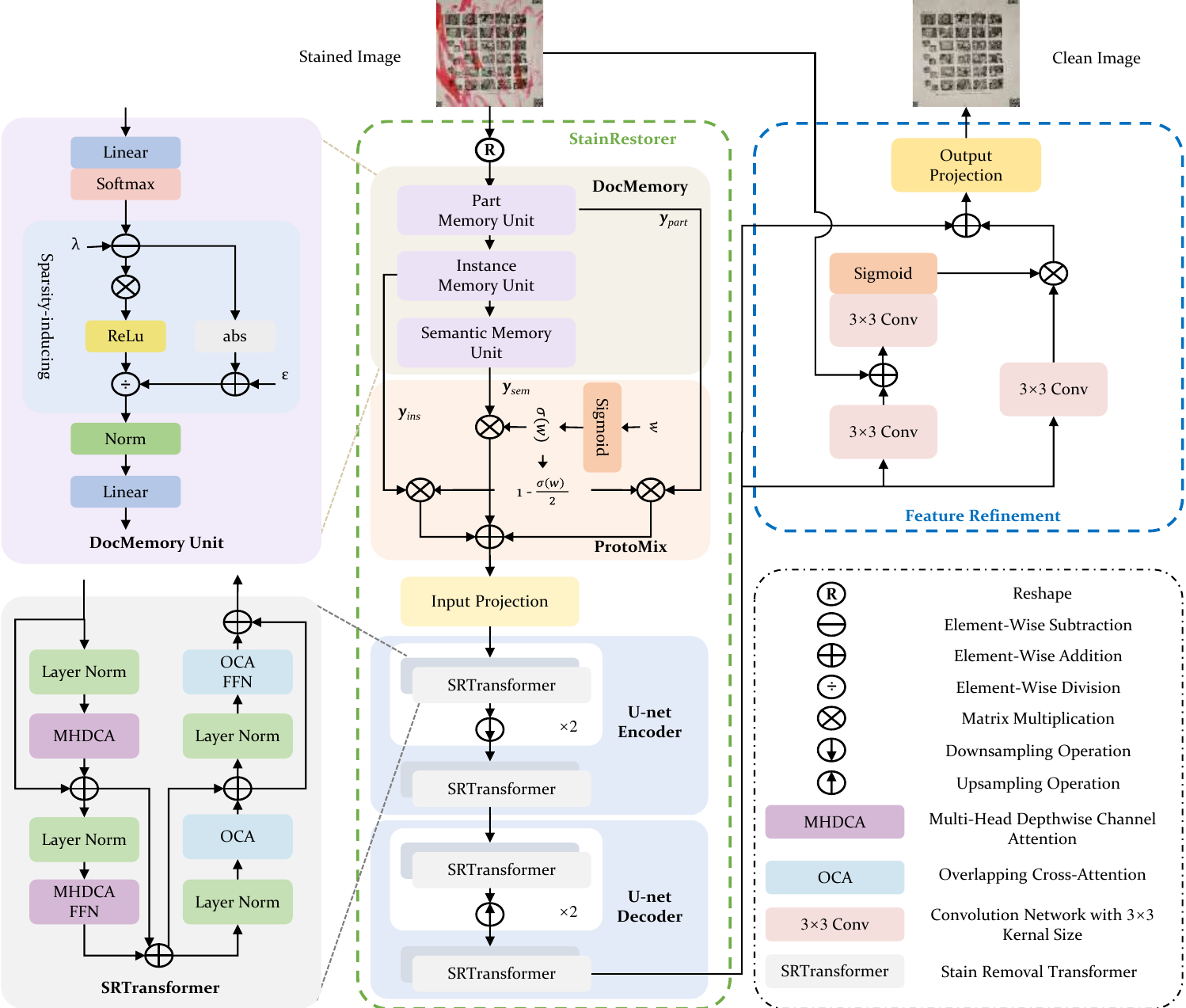}}
    \end{minipage}
    \caption{
    The overall architecture of the proposed StainRestorer. 
    It consists of a DocMemory module for hierarchical stain representation and a Stain Removal Transformer (SRTransformer) for accurate stain removal. The DocMemory module captures part-level, instance-level, and semantic-level stain features, which are then fused using the ProtoMix strategy. The SRTransformer leverages these rich representations to perform precise stain removal while preserving document content.
    }
    \label{fig:model}
\end{figure*}
\subsection{Overview}
StainRestorer, illustrated in~\cref{fig:model}, distinguishes itself through its hierarchical approach to stain representation and removal. The model first leverages the DocMemory module to capture diverse stain representations at multiple levels: part, instance, and semantic. This hierarchical representation facilitates a comprehensive understanding of the stain characteristics. Subsequently, the SRTransformer module, inspired by Restormer~\cite{chen2023comparative}, utilizes these rich representations to perform accurate stain removal. By combining the hierarchical representation learning of DocMemory with the powerful spatial mapping capabilities of SRTransformer, StainRestorer achieves high-quality document stain removal, effectively preserving the underlying document content.

\subsection{DocMemory Module}
The DocMemory module serves as the foundation of StainRestorer, responsible for extracting and analyzing deep features at various levels of granularity within the document. This module employs a three-level hierarchical structure:
\begin{itemize}
    \item Part-Level: Captures fine-grained details and local patterns within the stain.
    \item Instance-Level: Aggregates information from part-level features to represent individual stain instances.
    \item Semantic-Level: Extracts high-level semantic information and relationships between different stain instances.
\end{itemize}
This hierarchical design enables StainRestorer to progressively refine its understanding of the document's content and the characteristics of the stains present.

The DocMemory module achieves this hierarchical representation through a series of stacked Memory Units. Each unit operates as a library of learned visual prototypes, analogous to a visual dictionary. These prototypes represent characteristic visual patterns within the document. The core objective of each Memory Unit is to identify the prototypes that resonate most strongly with the input features, effectively quantifying the presence of specific visual patterns at a particular level of granularity.

Formally, the memory bank within each unit is represented by a matrix $\mathbf{M} \in \mathbb{R}^{N \times C}$, where $N$ denotes the number of memory items (prototypes) and $C$ represents the fixed feature dimension. The number of prototypes, $N$, is a hyperparameter adjustable based on the complexity of the task and the dataset.
The relevance of each prototype to a given input feature $\mathbf{f}_i$ is determined by calculating a similarity score using cosine similarity:
\begin{equation}
\begin{aligned}
s_{ij} &= \frac{\exp(d(\mathbf{f}_i,\mathbf{m}_j))}{\sum_{j=1}^{M} \exp(d(\mathbf{f}_i,\mathbf{m}_j))},\\
d\left(\mathbf{f}_{i},\mathbf{m}_{j}\right) &= \frac{\mathbf{f}_{i}\mathbf{m}_{j}^{\top}}{\left\|\mathbf{f}_{i}\right\|\left\|\mathbf{m}_{j}\right\|},
\end{aligned}
\end{equation}
where $\mathbf{m}j$ represents the $j$-th prototype in the memory bank. The resulting attention weight $s_{ij}$ quantifies the relevance of the $j$-th prototype to the input feature vector $\mathbf{f}_i$.

To further enhance the efficiency and discriminative capability of the memory access, a sparsity-inducing mechanism is introduced. This mechanism operates on the attention weights, promoting a selective focus on the most informative prototypes by suppressing activations below a certain threshold.

The final output of a Memory Unit is a weighted sum of the prototypes:
\begin{equation}
\mathbf{y} = \operatorname{Memory}(\mathbf{f},\mathbf{M}) = \sum_{j=1}^K {s}_{ij} \mathbf{m}_j.
\end{equation}
The hierarchical nature of DocMemory is reflected in the way higher-level prototypes are derived. Each higher-level prototype summarizes information from its lower-level counterparts. Specifically:
\begin{equation}
\begin{aligned}
\mathbf{y}_{\mathrm{part}} &= M_{\mathrm{part}} (\mathbf{f},\mathbf{M}_{\mathrm{part}} ),\\
\mathbf{y}_{\mathrm{ins}} &= M_{\mathrm{ins}} (\mathbf{y}_{\mathrm{part}} ,\mathbf{M}_{\mathrm{ins}} ),\\
\mathbf{y}_{\mathrm{sem}} &= M_{\mathrm{part}} (\mathbf{y}_{\mathrm{ins}} ,\mathbf{M}_{\mathrm{sem}} ).
\end{aligned}
\end{equation}
This hierarchical workflow ensures that the final representation captures both local details and global context.

Finally, to leverage the complementary information embedded within these distinct feature hierarchies, a learnable mixing mechanism called ProtoMix dynamically combines the part-level, instance-level, and semantic-level features. This allows the model to adaptively emphasize different levels of representation based on the specific task requirements:
\begin{equation}
\mathbf{y}_{\mathrm{mix}} = \sigma(w) \cdot \mathbf{y}_{\mathrm{sem}} + \frac{1 - \sigma(w)}{2} \cdot \mathbf{y}_{\mathrm{ins}} + \frac{1 - \sigma(w)}{2} \cdot \mathbf{y}_{\mathrm{part}},
\end{equation}
where $\sigma(w)$ represents the sigmoid function applied to the learnable weight parameter $w$. This mixing strategy empowers the model to effectively capture both fine-grained details and high-level semantic information, leading to a more holistic and informative document representation.

\subsection{Stain Removal Transformer (SRTransformer)}
Inspired by the effectiveness of Restormer~\cite{chen2023comparative} in image restoration tasks, StainRestorer employs a modified Transformer architecture named Stain Removal Transformer (SRTransformer). This module is specifically designed to enhance the model's spatial mapping ability for precise stain removal.

SRTransformer incorporates two complementary attention mechanisms. Multi-Head Depthwise Channel Attention (MHDCA) focuses on channel-wise relationships by applying $1\times1$ convolution and $3\times3$ depthwise convolutions to generate  query ($Q$), key ($K$), and value ($V$) tensors, which are then split into multiple heads, normalized, and used to compute attention weights with a learnable temperature parameter. The output is reshaped and passed through a $1\times1$ convolution, emphasizing crucial channels for stain removal. Overlapping Cross-Attention (OCA)~\cite{10204527}: Enhances spatial attention by expanding the receptive field for key ($K$) and value ($V$) computations. This enables each query ($Q$) to access a broader context, improving the model's ability to capture long-range dependencies and spatial relationships relevant to stain removal.

\begin{figure*}[ht]
    \begin{minipage}[b]{1.0\linewidth}
        \begin{minipage}[b]{0.12\linewidth}
            \centering
            \centerline{\includegraphics[width=\linewidth]{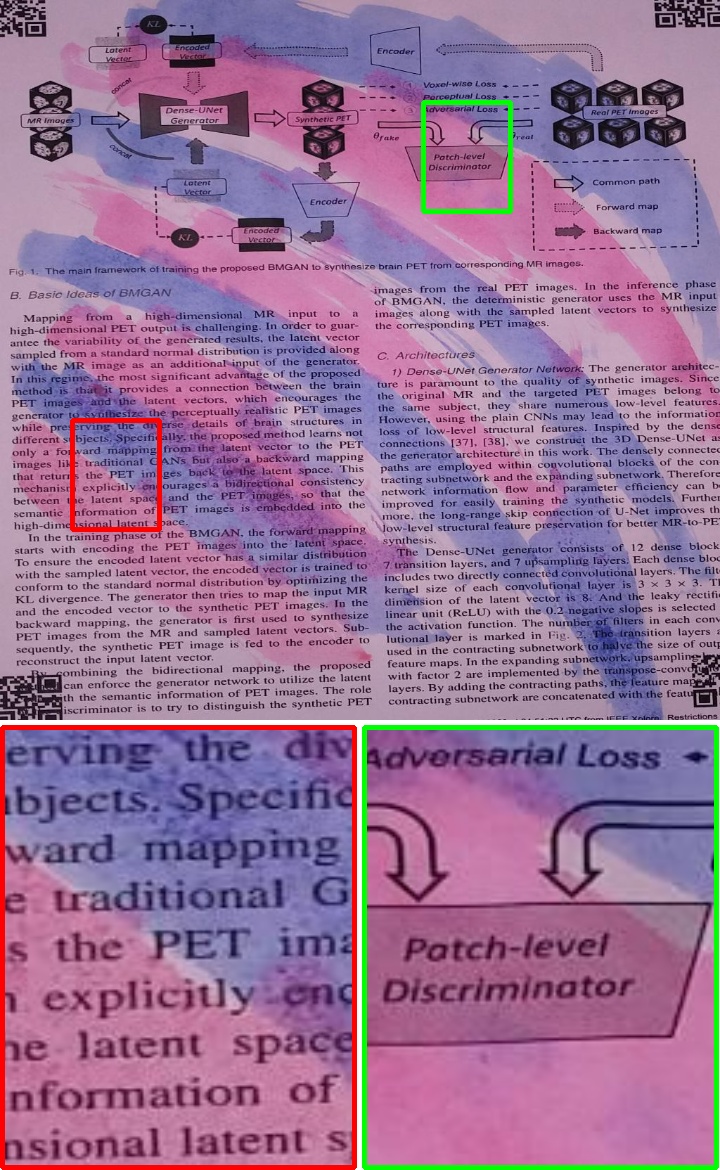}}
        \end{minipage}
        \hfill
        \begin{minipage}[b]{0.12\linewidth}
            \centering
            \centerline{\includegraphics[width=\linewidth]{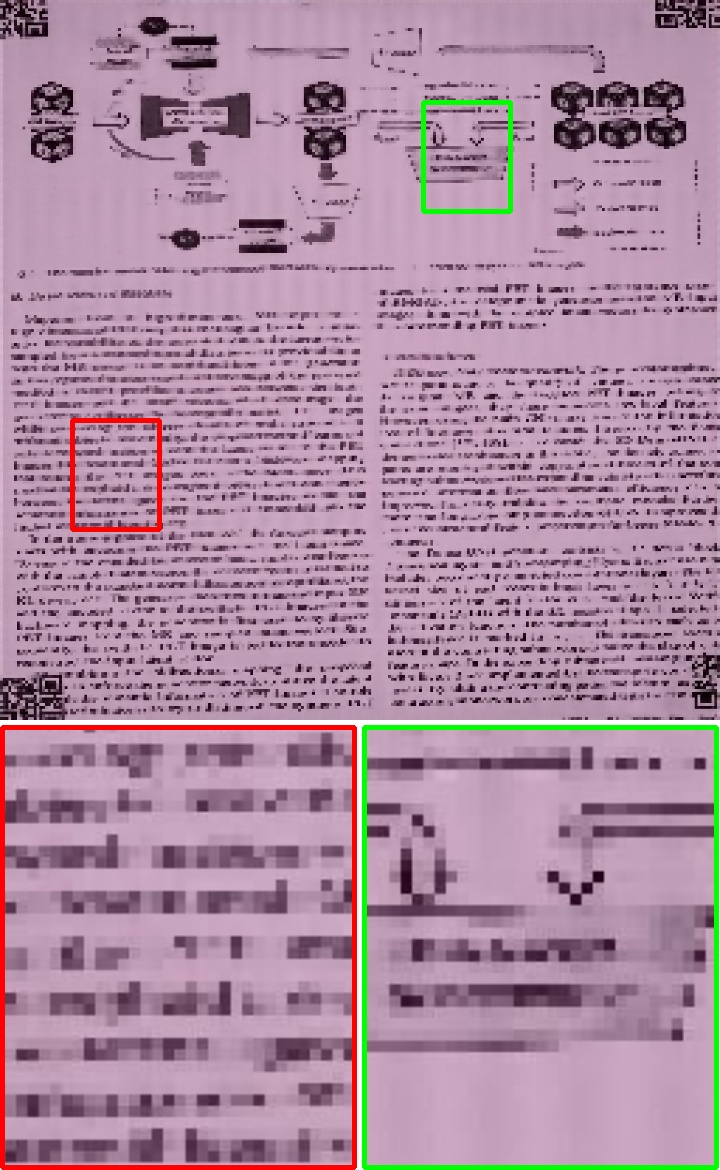}}
        \end{minipage}
        \hfill
        \begin{minipage}[b]{0.12\linewidth}
            \centering
            \centerline{\includegraphics[width=\linewidth]{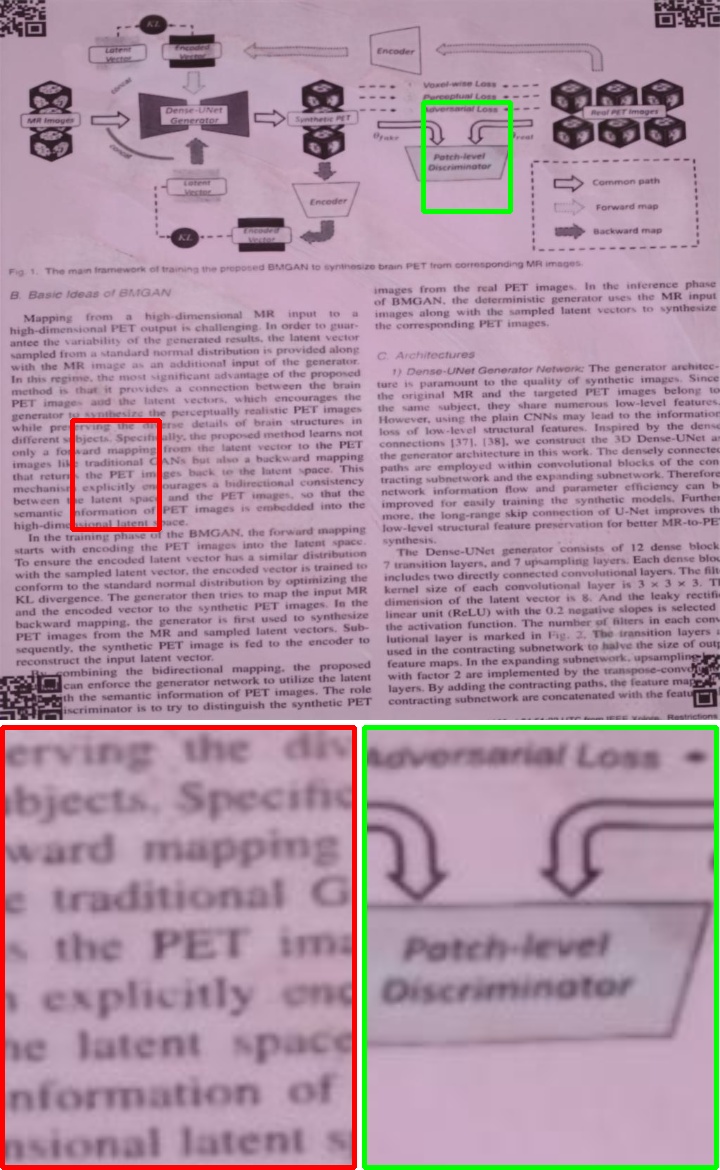}}
        \end{minipage}
        \hfill
        \begin{minipage}[b]{0.12\linewidth}
            \centering
            \centerline{\includegraphics[width=\linewidth]{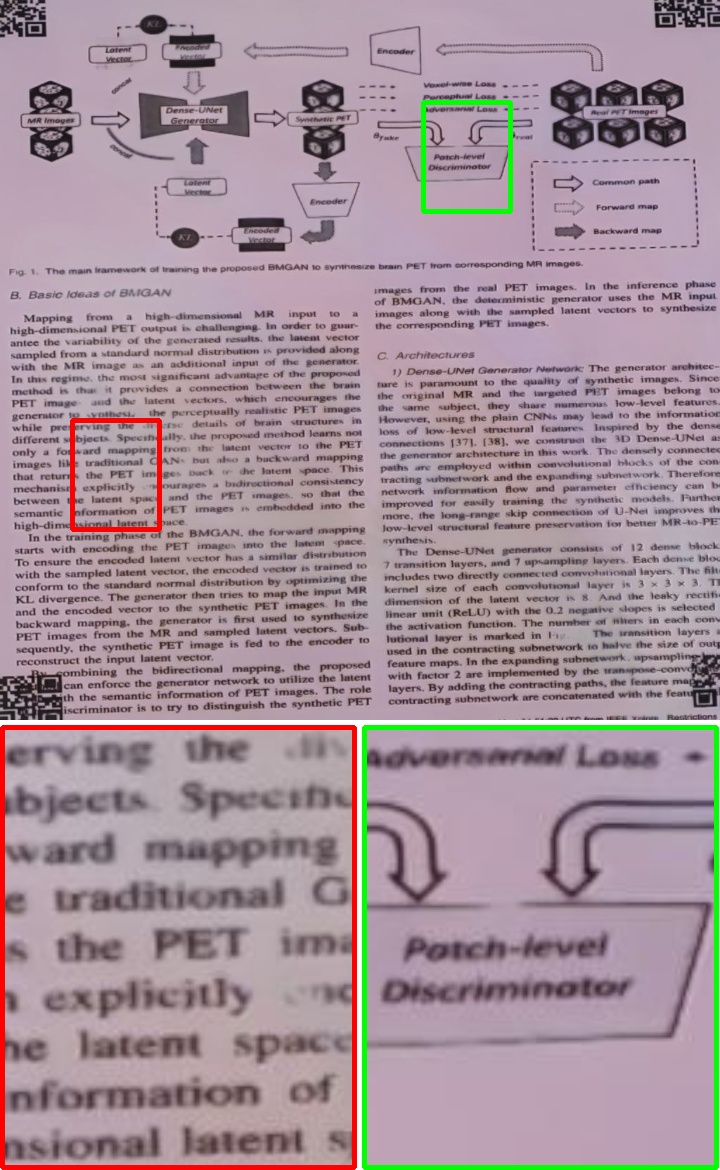}}
        \end{minipage}
        \hfill
        \begin{minipage}[b]{0.12\linewidth}
            \centering
            \centerline{\includegraphics[width=\linewidth]{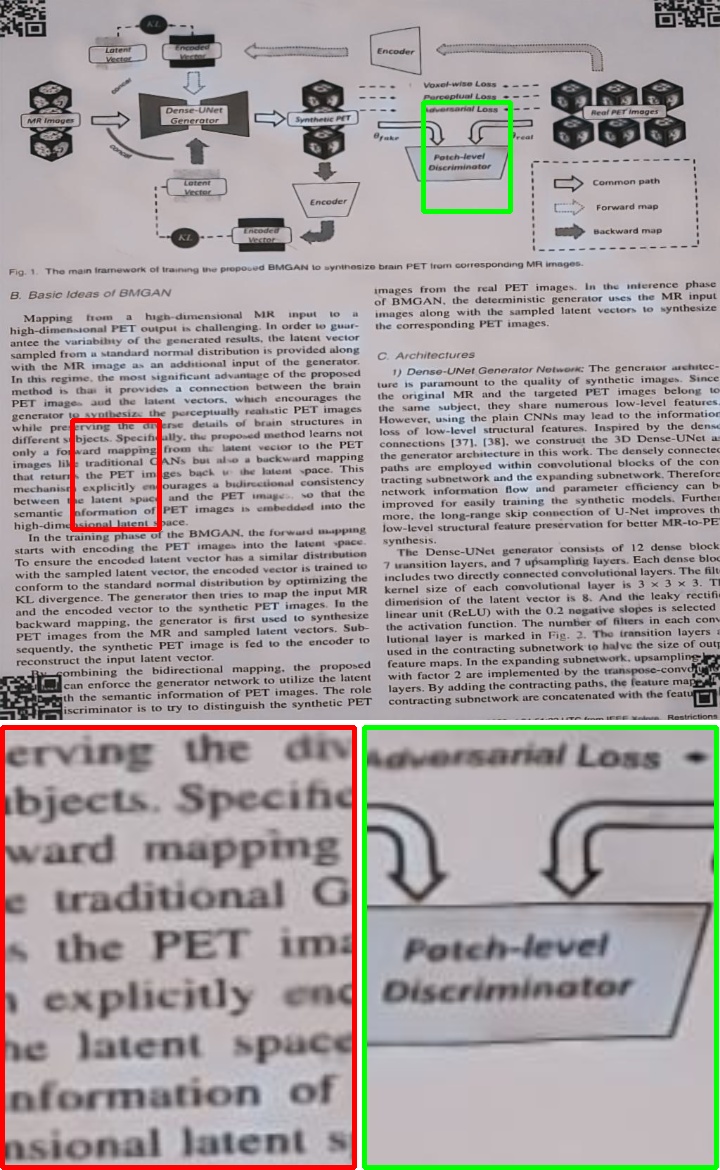}}
        \end{minipage}
        \hfill
        \begin{minipage}[b]{0.12\linewidth}
            \centering
            \centerline{\includegraphics[width=\linewidth]{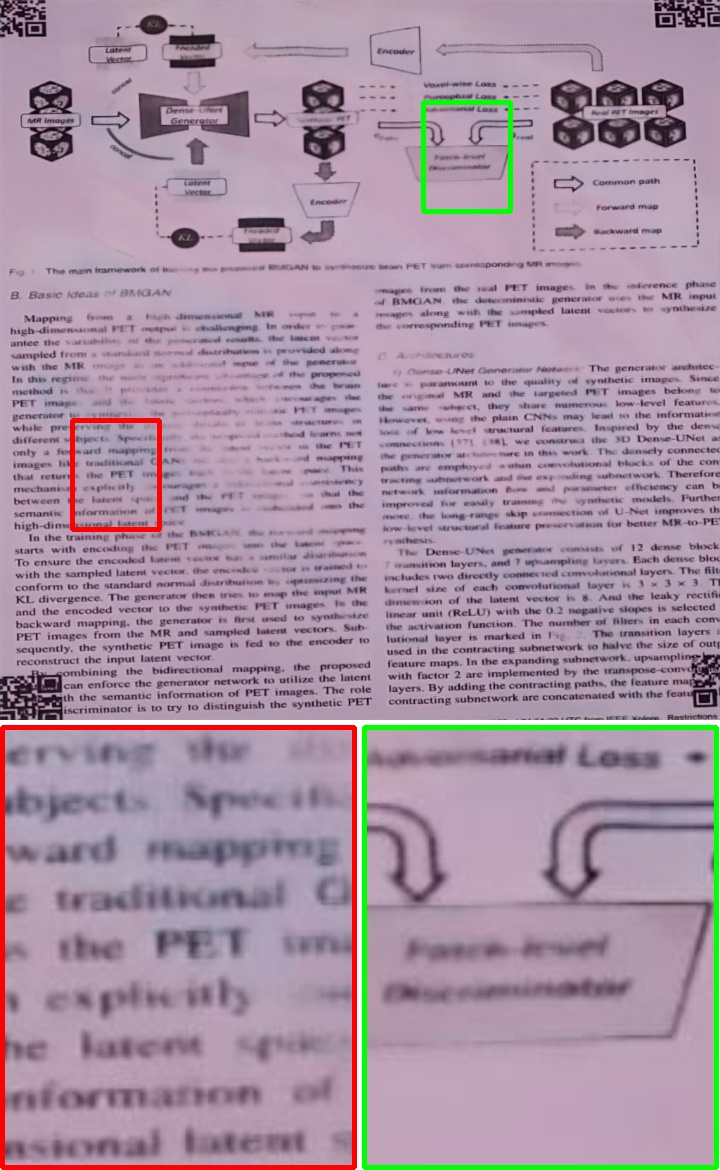}}
        \end{minipage}
        \hfill
        \begin{minipage}[b]{0.12\linewidth}
            \centering
            \centerline{\includegraphics[width=\linewidth]{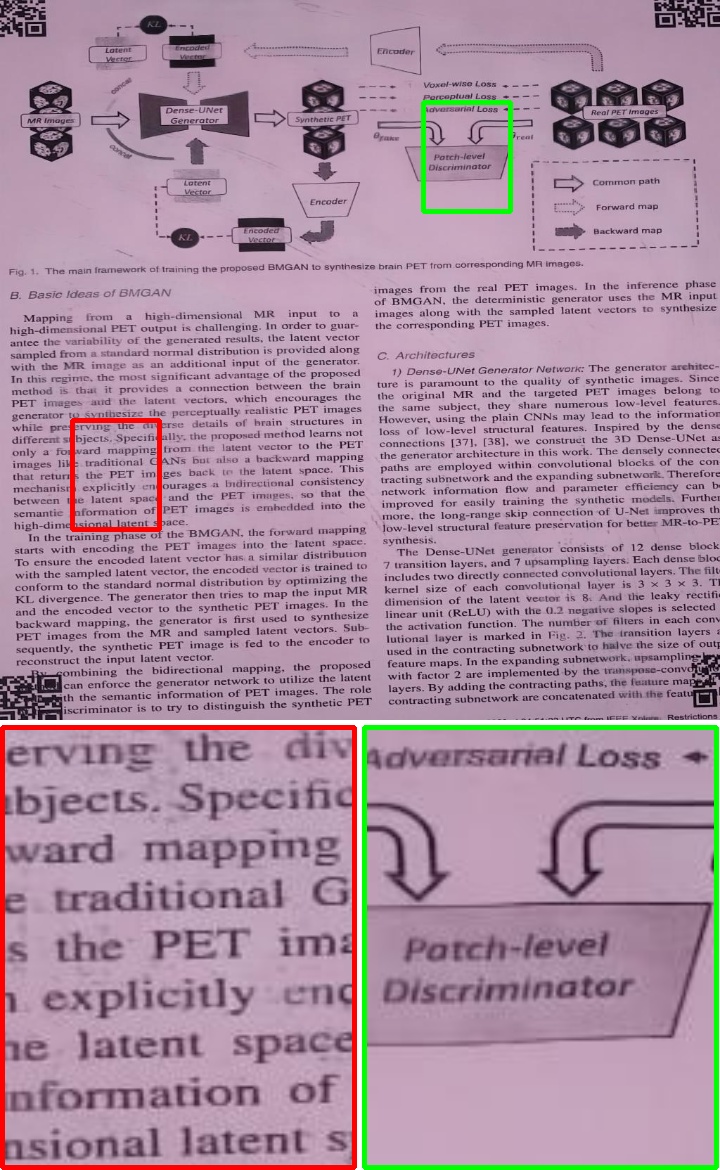}}
        \end{minipage}
        \hfill
        \begin{minipage}[b]{0.12\linewidth}
            \centering
            \centerline{\includegraphics[width=\linewidth]{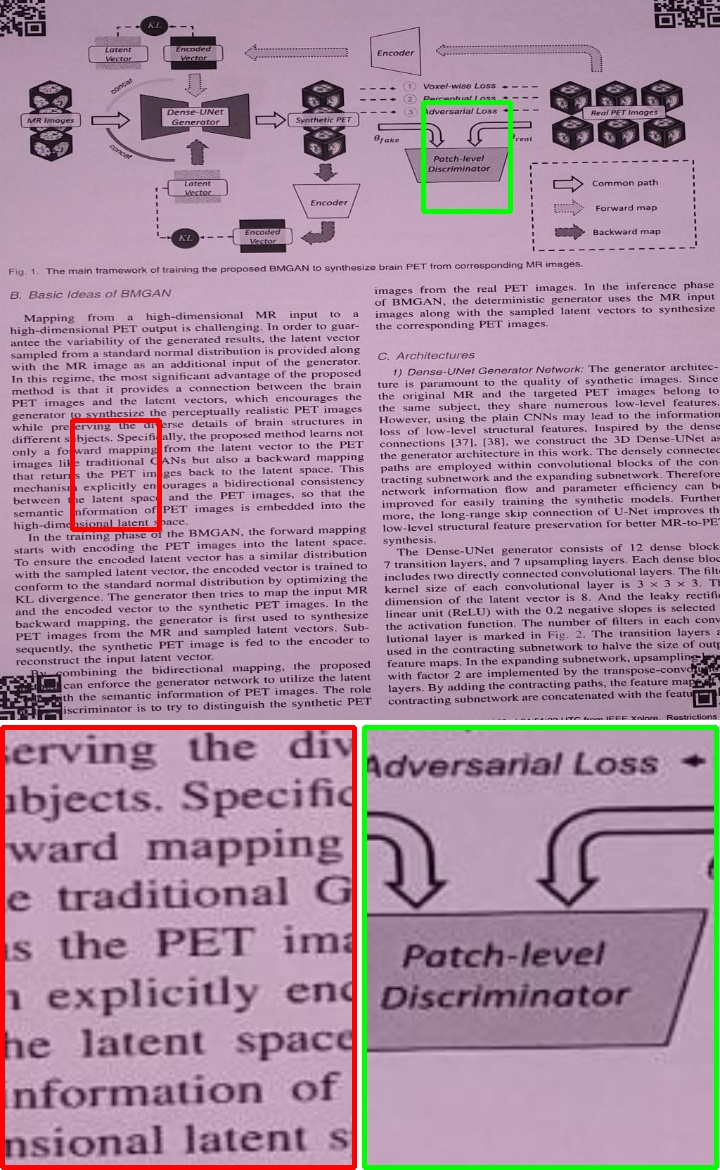}}
        \end{minipage}

    \end{minipage}
    \begin{minipage}[b]{1.0\linewidth}
        \begin{minipage}[b]{0.12\linewidth}
            \centering
            \centerline{\includegraphics[width=\linewidth]{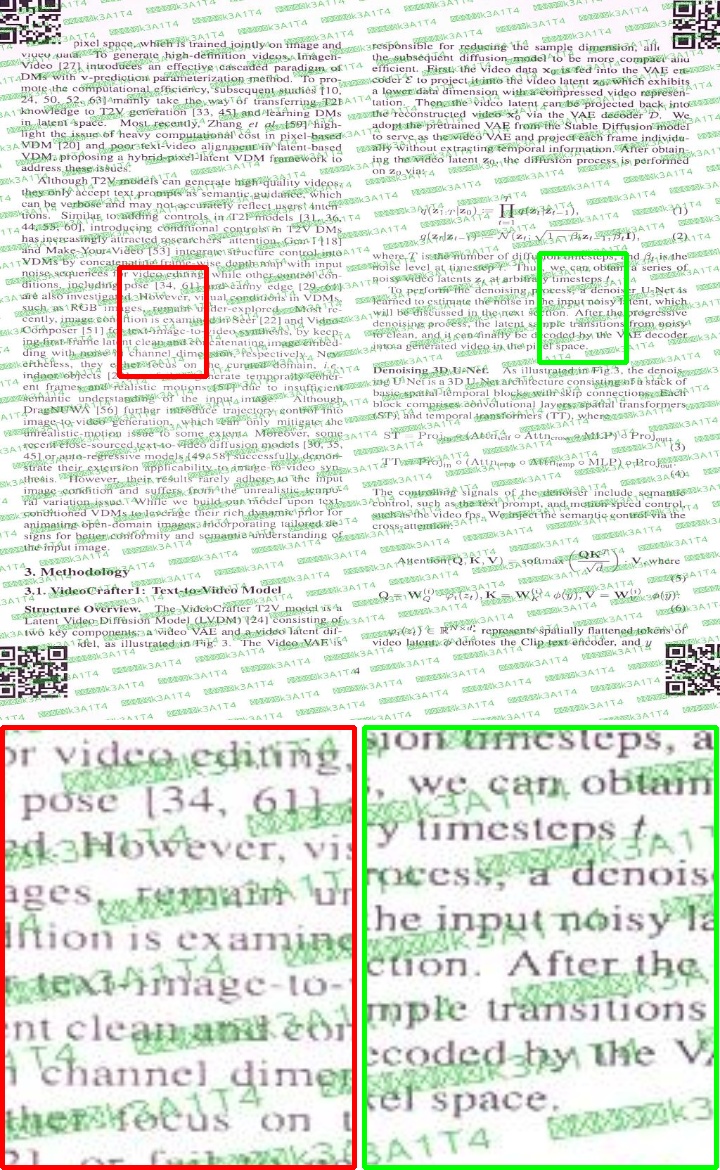}}
        \end{minipage}
        \hfill
        \begin{minipage}[b]{0.12\linewidth}
            \centering
            \centerline{\includegraphics[width=\linewidth]{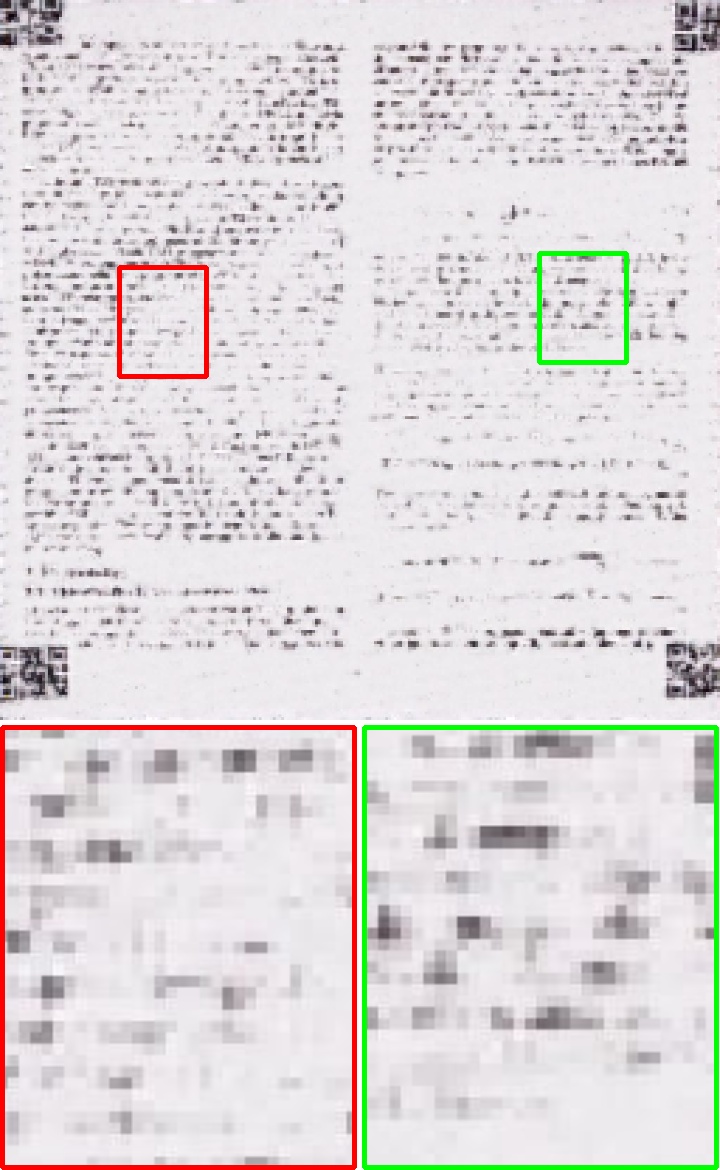}}
        \end{minipage}
        \hfill
        \begin{minipage}[b]{0.12\linewidth}
            \centering
            \centerline{\includegraphics[width=\linewidth]{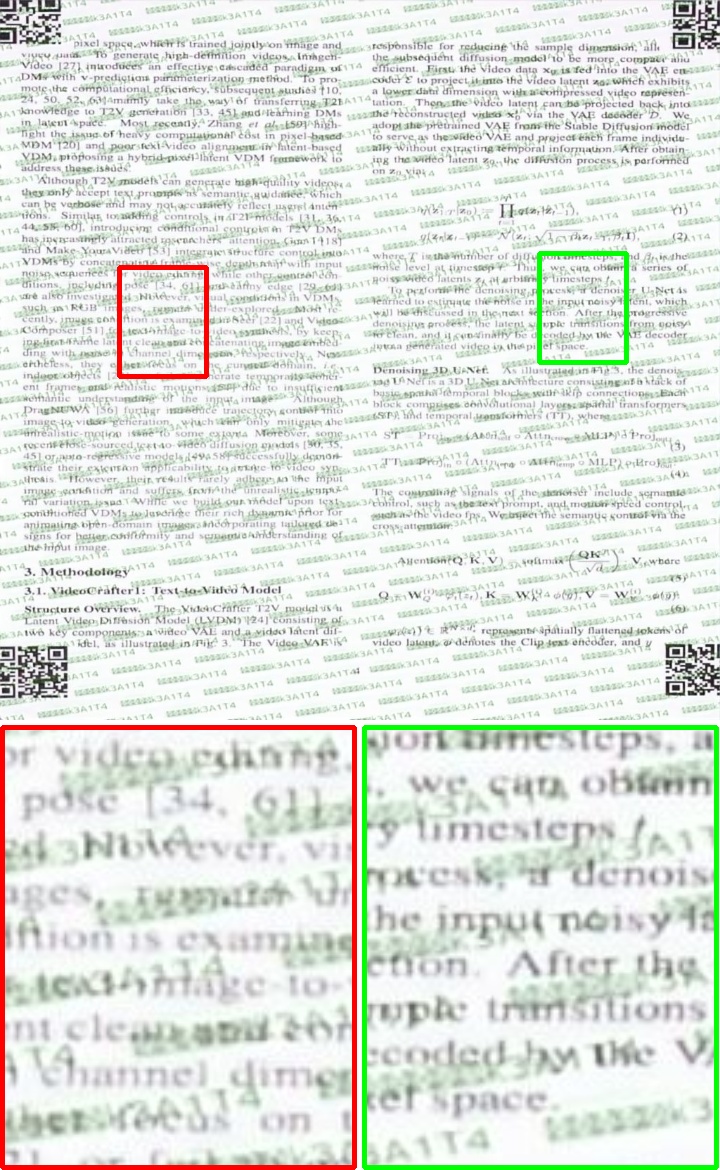}}
        \end{minipage}
        \hfill
        \begin{minipage}[b]{0.12\linewidth}
            \centering
            \centerline{\includegraphics[width=\linewidth]{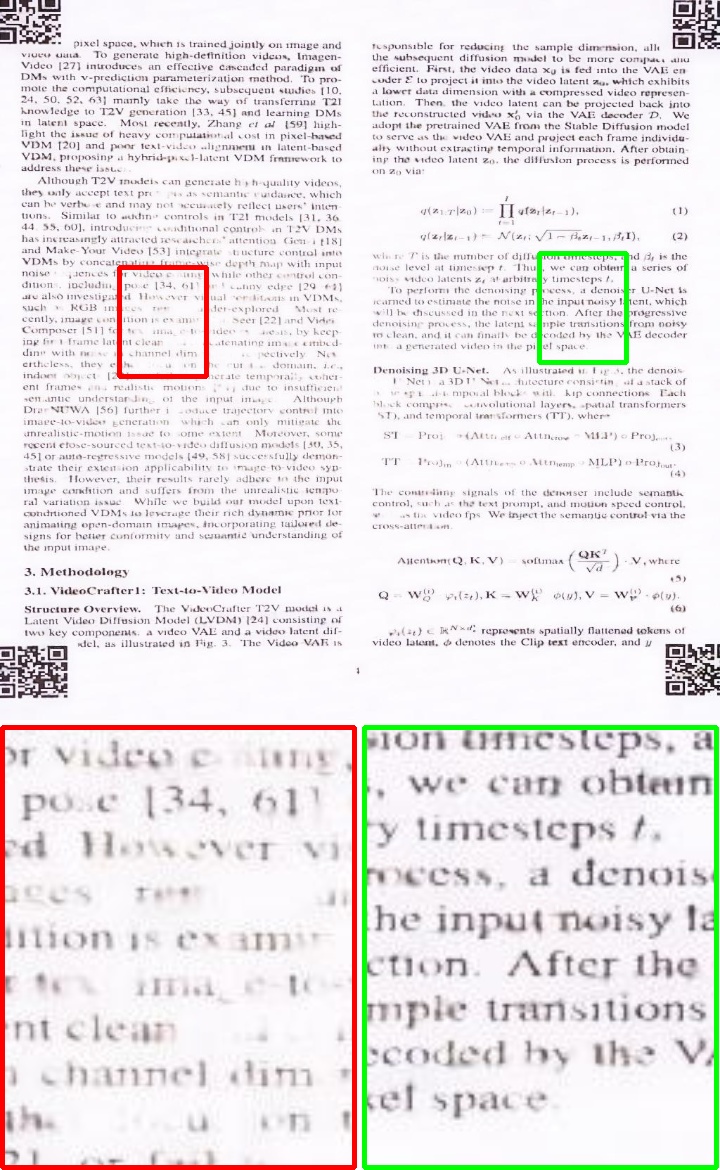}}
        \end{minipage}
        \hfill
        \begin{minipage}[b]{0.12\linewidth}
            \centering
            \centerline{\includegraphics[width=\linewidth]{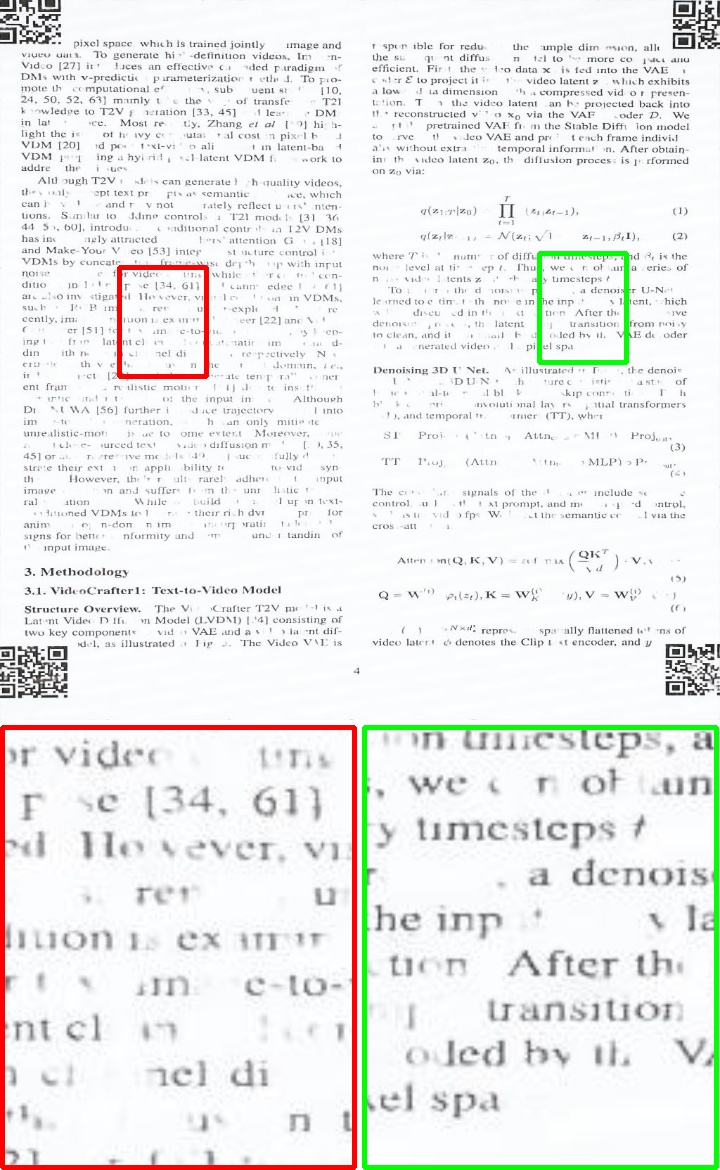}}
        \end{minipage}
        \hfill
        \begin{minipage}[b]{0.12\linewidth}
            \centering
            \centerline{\includegraphics[width=\linewidth]{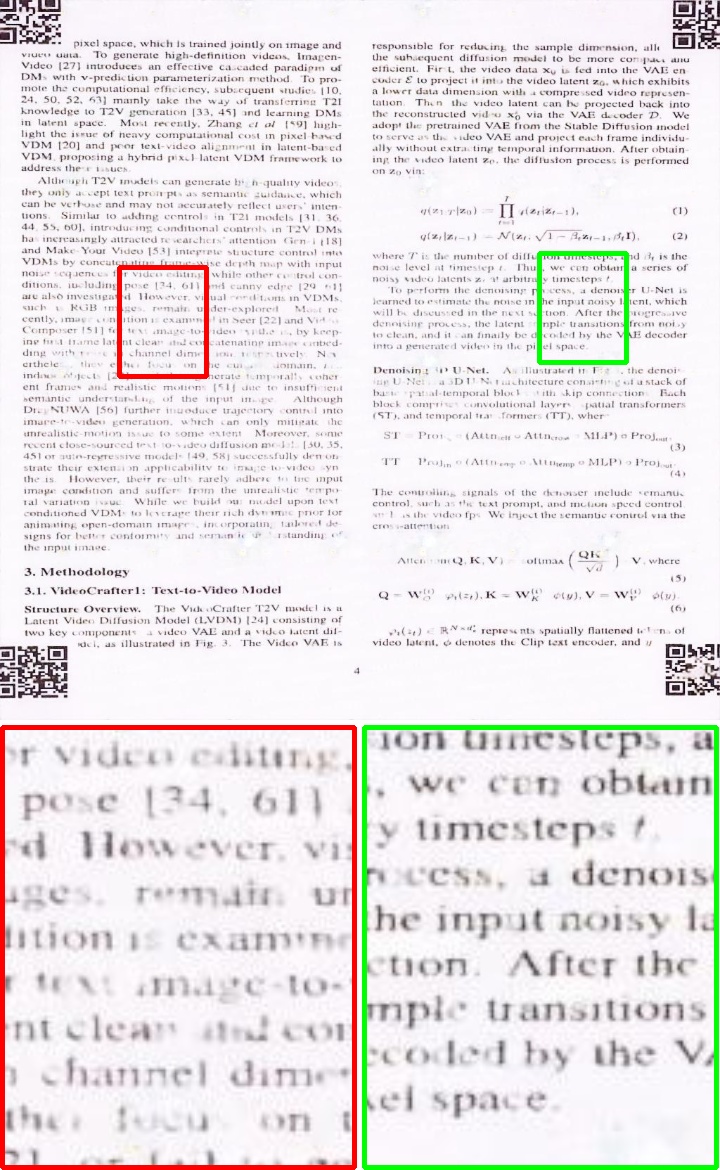}}
        \end{minipage}
        \hfill
        \begin{minipage}[b]{0.12\linewidth}
            \centering
            \centerline{\includegraphics[width=\linewidth]{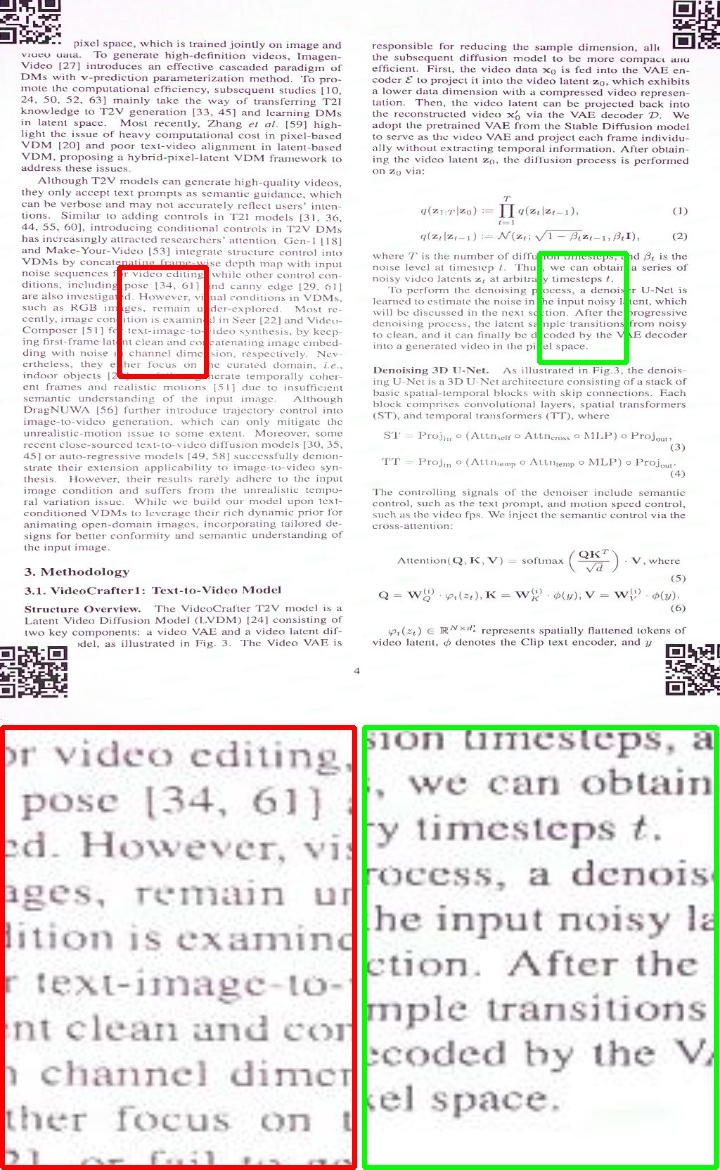}}
        \end{minipage}
        \hfill
        \begin{minipage}[b]{0.12\linewidth}
            \centering
            \centerline{\includegraphics[width=\linewidth]{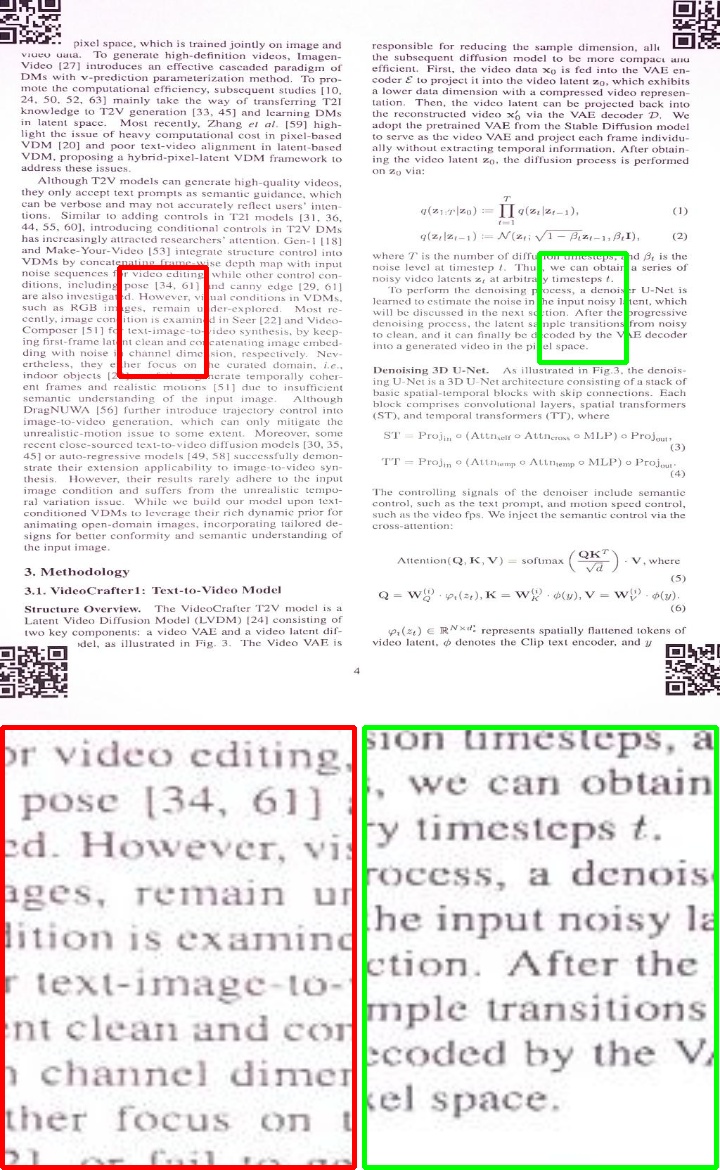}}
        \end{minipage}

    \end{minipage}

    \begin{minipage}[b]{1.0\linewidth}
        \begin{minipage}[b]{0.12\linewidth}
            \centering
            \centerline{\includegraphics[width=\linewidth]{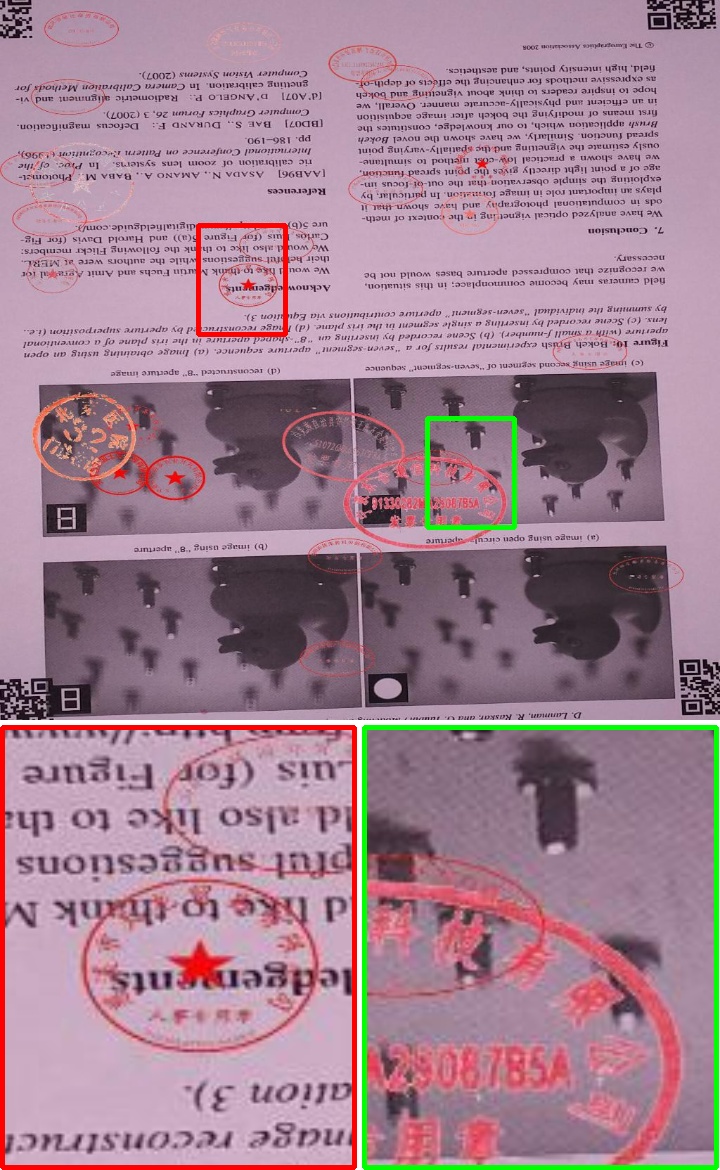}}
            \centerline{(a) Input}\medskip
        \end{minipage}
        \hfill
        \begin{minipage}[b]{0.12\linewidth}
            \centering
            \centerline{\includegraphics[width=\linewidth]{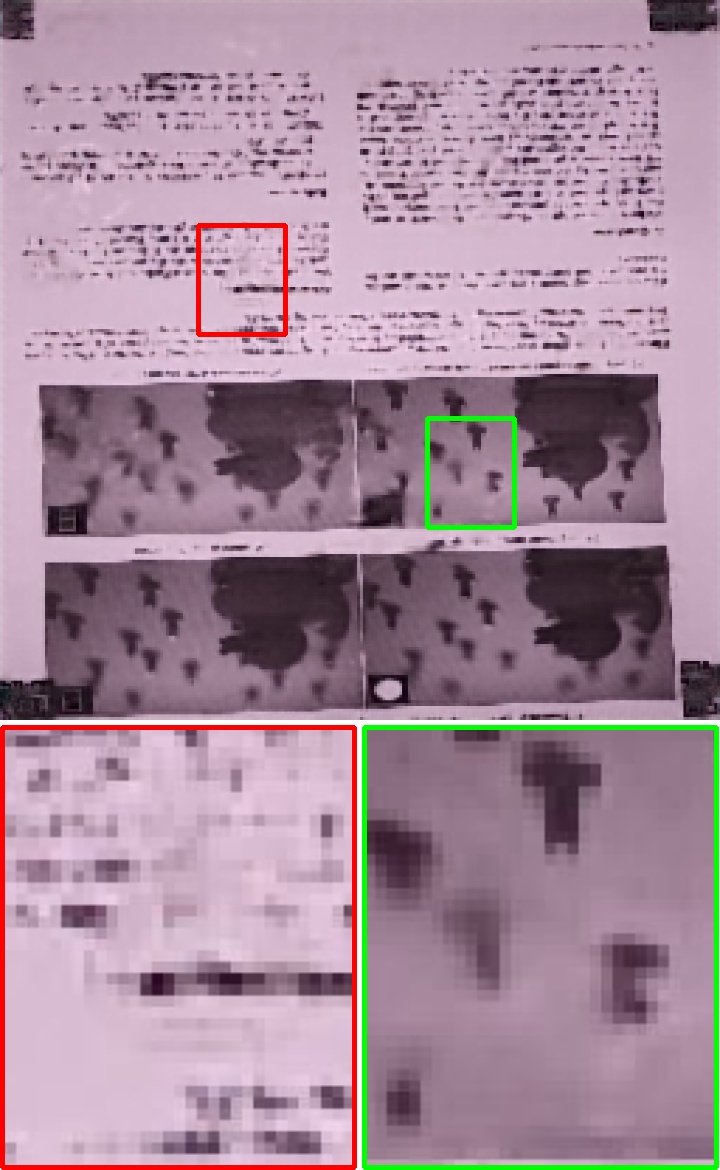}}
            \centerline{(b) illtrtemplate}\medskip
        \end{minipage}
        \hfill
        \begin{minipage}[b]{0.12\linewidth}
            \centering
            \centerline{\includegraphics[width=\linewidth]{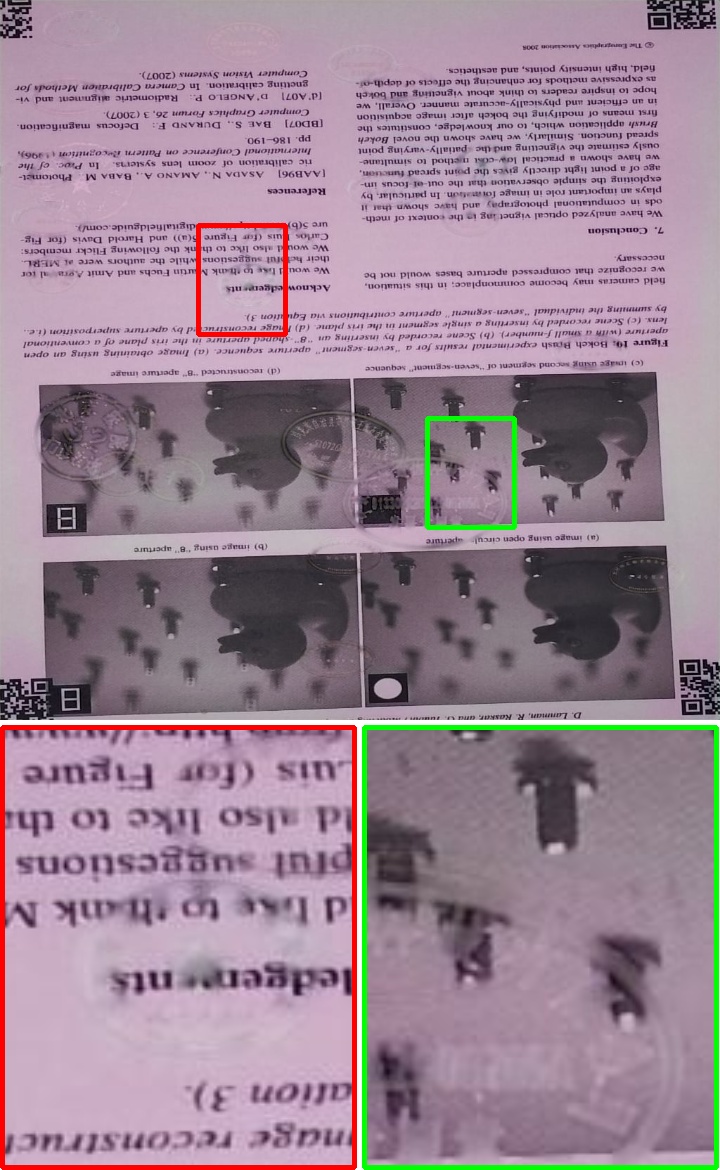}}
            \centerline{(c) DocNLC}\medskip
        \end{minipage}
        \hfill
        \begin{minipage}[b]{0.12\linewidth}
            \centering
            \centerline{\includegraphics[width=\linewidth]{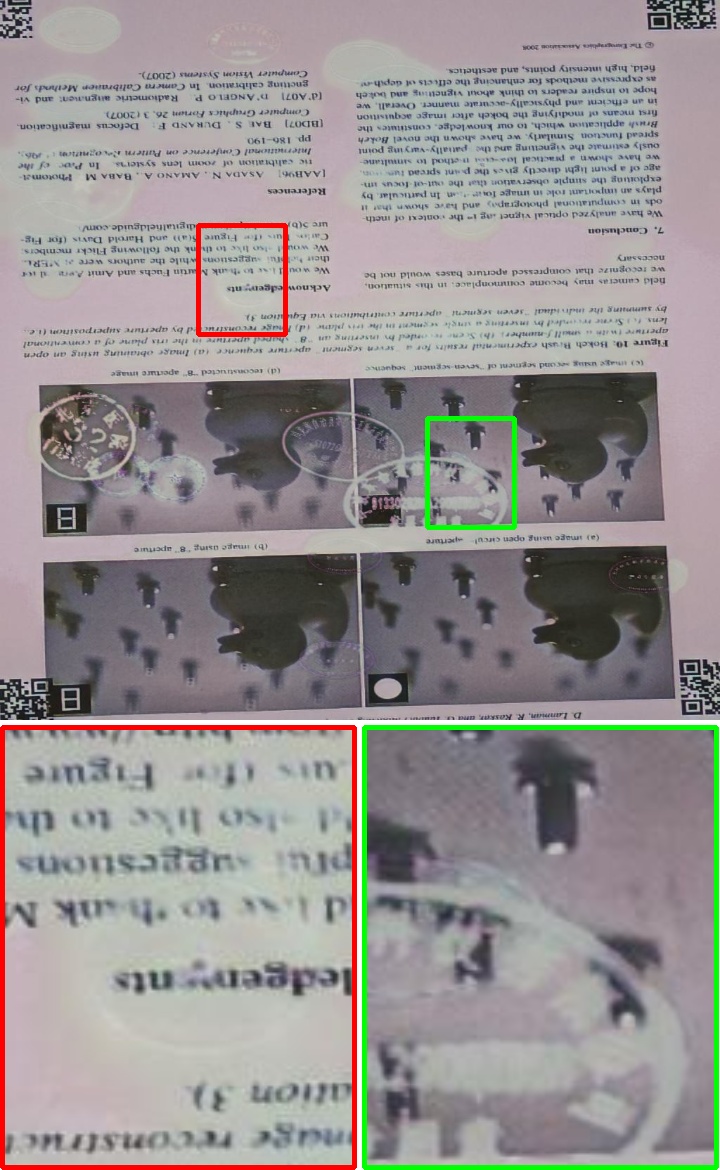}}
            \centerline{(d) GCDRNet}\medskip
        \end{minipage}
        \hfill
        \begin{minipage}[b]{0.12\linewidth}
            \centering
            \centerline{\includegraphics[width=\linewidth]{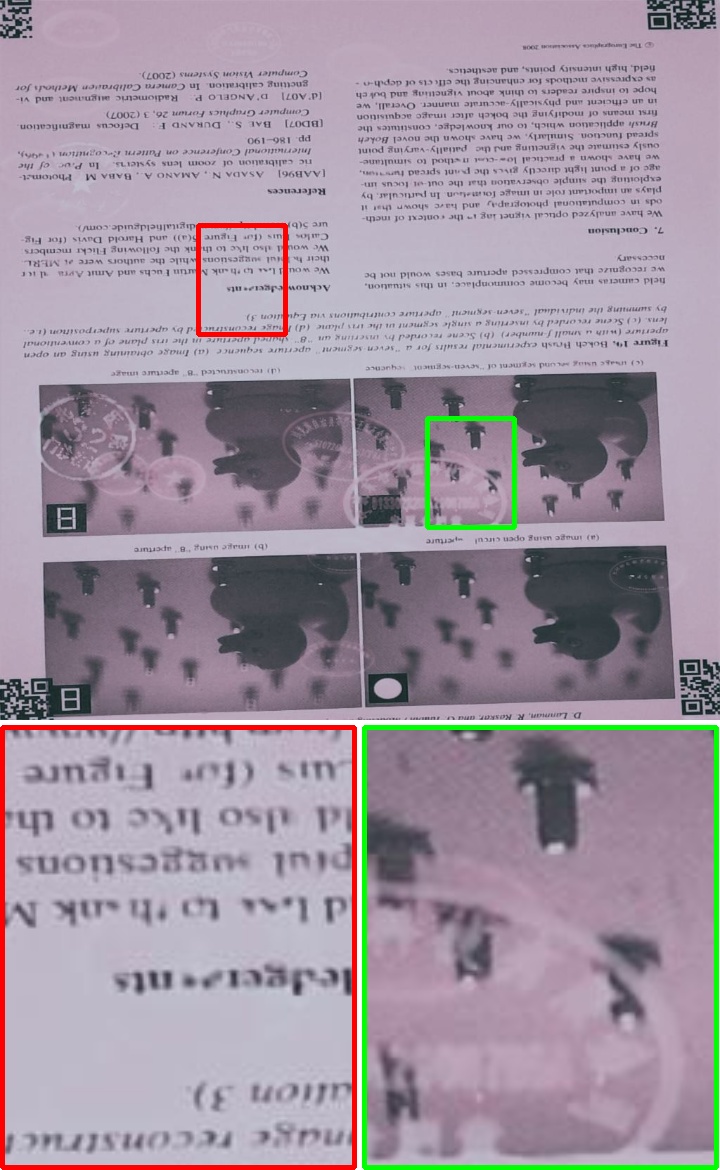}}
            \centerline{(e) DEGAN}\medskip
        \end{minipage}
        \hfill
        \begin{minipage}[b]{0.12\linewidth}
            \centering
            \centerline{\includegraphics[width=\linewidth]{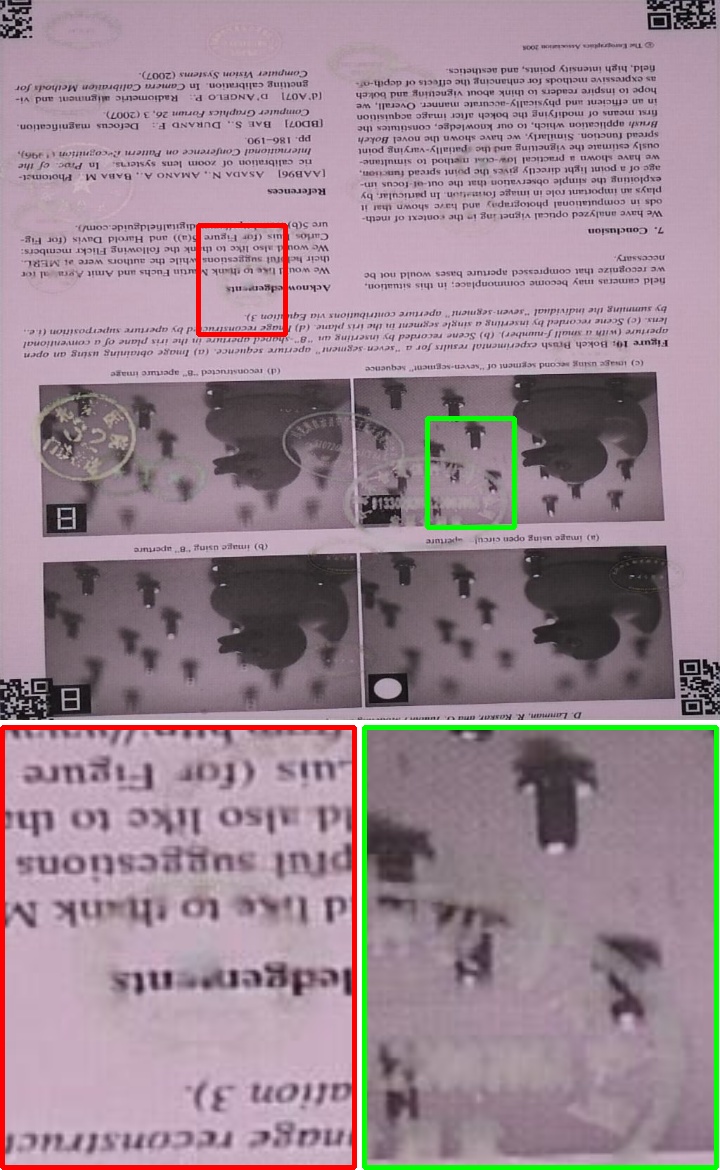}}
            \centerline{(f) UDoc-GAN}\medskip
        \end{minipage}
        \hfill
        \begin{minipage}[b]{0.12\linewidth}
            \centering
            \centerline{\includegraphics[width=\linewidth]{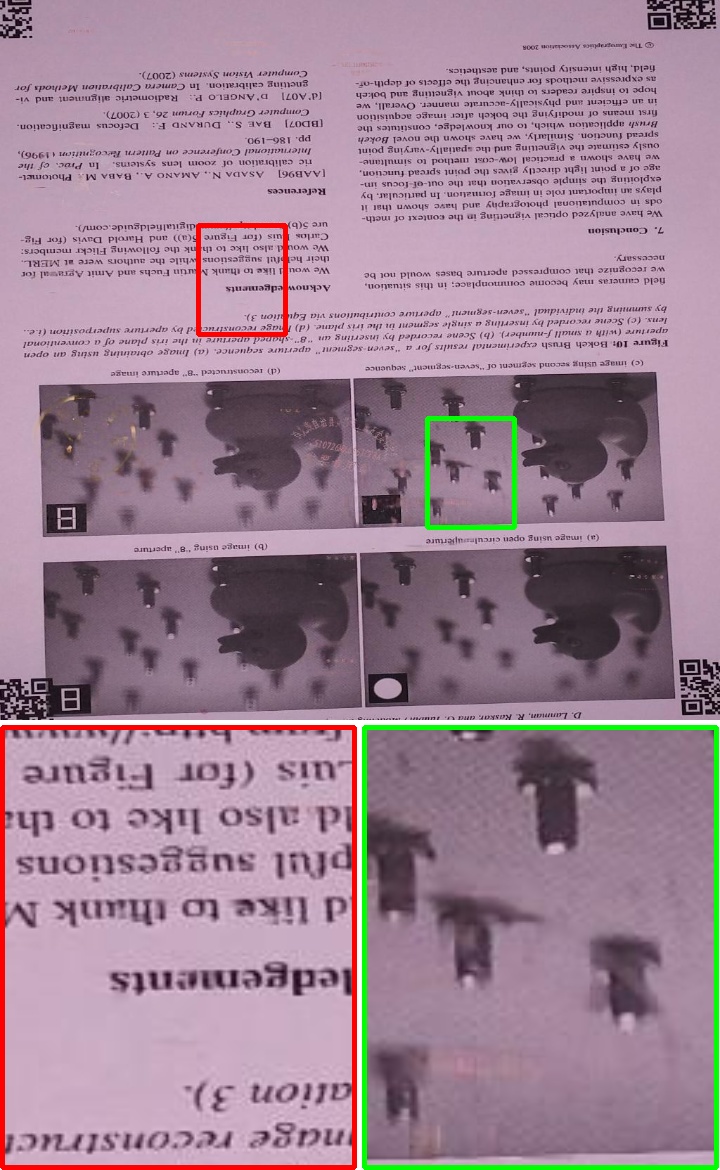}}
            \centerline{(g) Ours}\medskip
        \end{minipage}
        \hfill
        \begin{minipage}[b]{0.12\linewidth}
            \centering
            \centerline{\includegraphics[width=\linewidth]{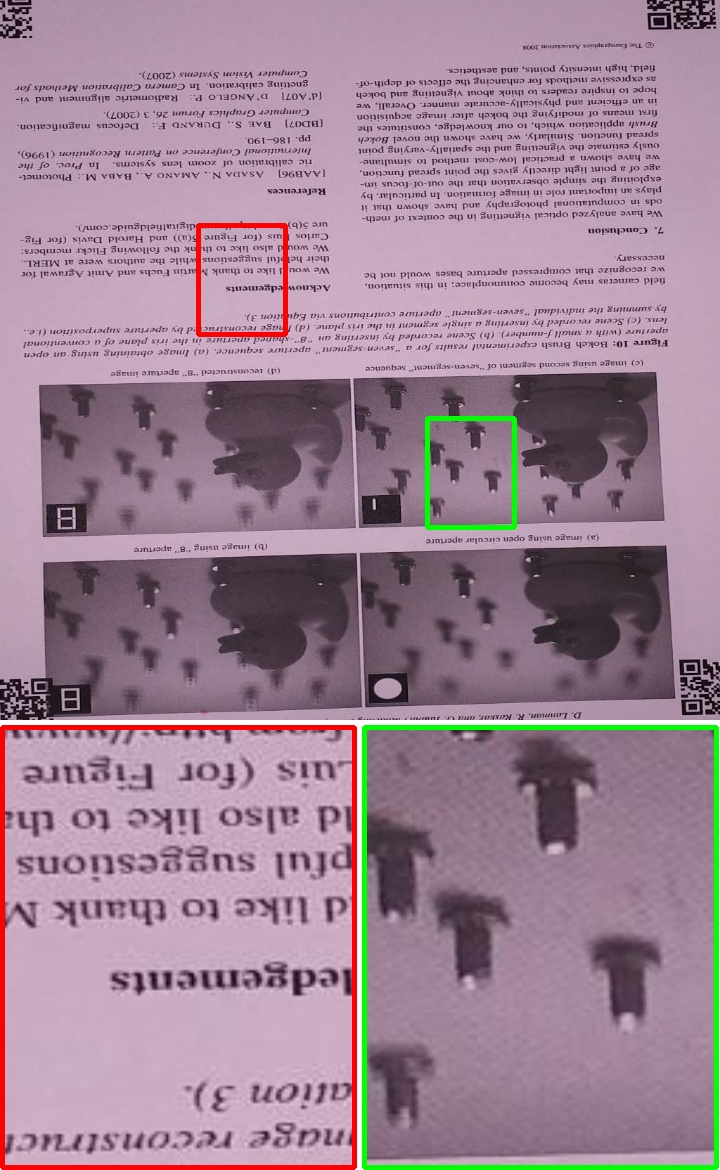}}
            \centerline{(h) Target}\medskip
        \end{minipage}

    \end{minipage}

    \caption{Qualitative comparison of stain removal performance across different models. 
    The first row presents results on the StainDoc dataset, the second row on the StainDoc\_Mark dataset, and the third row on the StainDoc\_Seal dataset.
    }
    \label{fig:compare}
    
\end{figure*}

The SRTransformer operates within a U-net encoder-decoder structure, as shown in~\cref{fig:model}. It takes the output of DocMemory's ProtoMix module, $\mathbf{y}_{\mathrm{mix}}$, and projects it into a higher-dimensional embedding space to enhance its representational capacity, resulting in $F_\mathrm{in}$. The workflow of SRTransformer can be represented as follows:
\begin{equation}
\begin{aligned}
F_\mathrm{c}&=F_\mathrm{in}+\operatorname{MHDCA}(\operatorname{LayerNorm}(F_\mathrm{in})), \\
F_\mathrm{c_{out}}&=F_\mathrm{t}+\operatorname{FFN}(\operatorname{LayerNorm}(F_\mathrm{t})), \\
F_\mathrm{s}&=F_\mathrm{c_{out}}+\operatorname{OCA}(\operatorname{LayerNorm}(F_\mathrm{t_{out}})), \\
F_\mathrm{s_{out}}&=F_\mathrm{s}+\operatorname{FFN}(\operatorname{LayerNorm}(F_\mathrm{s})),        
\end{aligned}        
\end{equation}
where $F_\mathrm{c}$ and $F_\mathrm{c_{out}}$ denote the transitional feature and the resultant output from the $\operatorname{MHDCA}$ respectively. $F_\mathrm{s}$ and $F_\mathrm{s_{out}}$ represent the intermediate feature and the final output of the $\operatorname{OCA}$. $\operatorname{MHDCA}$ symbolizes the Multi-Head Depthwise Channel Attention operation, while $\operatorname{OCA}$ represents the function of the Overlapping Cross-Attention. $\operatorname{LayerNorm}$ is used to indicate the Layer Normalization process. $\operatorname{FFN}$ stands for the Feed-Forward Network component of the architecture.

Through this process, SRTransformer refines the representation to distinguish between stain artifacts and genuine document details, leading to effective stain removal while preserving the integrity of the document content.
\subsection{Loss Function}
\begin{figure*}[ht]
    \begin{minipage}[b]{1.0\linewidth}
        \begin{minipage}[b]{0.16\linewidth}
            \centering
            \centerline{\includegraphics[width=\linewidth]{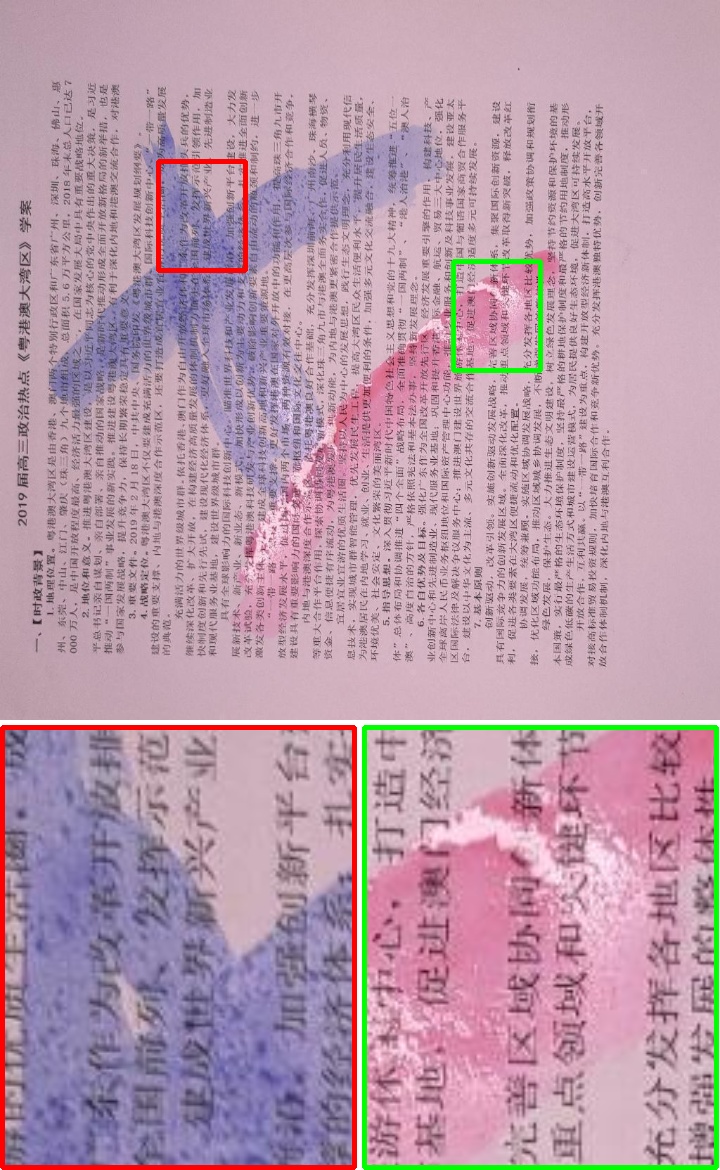}}
        \end{minipage}
        \hfill
        \begin{minipage}[b]{0.16\linewidth}
            \centering
            \centerline{\includegraphics[width=\linewidth]{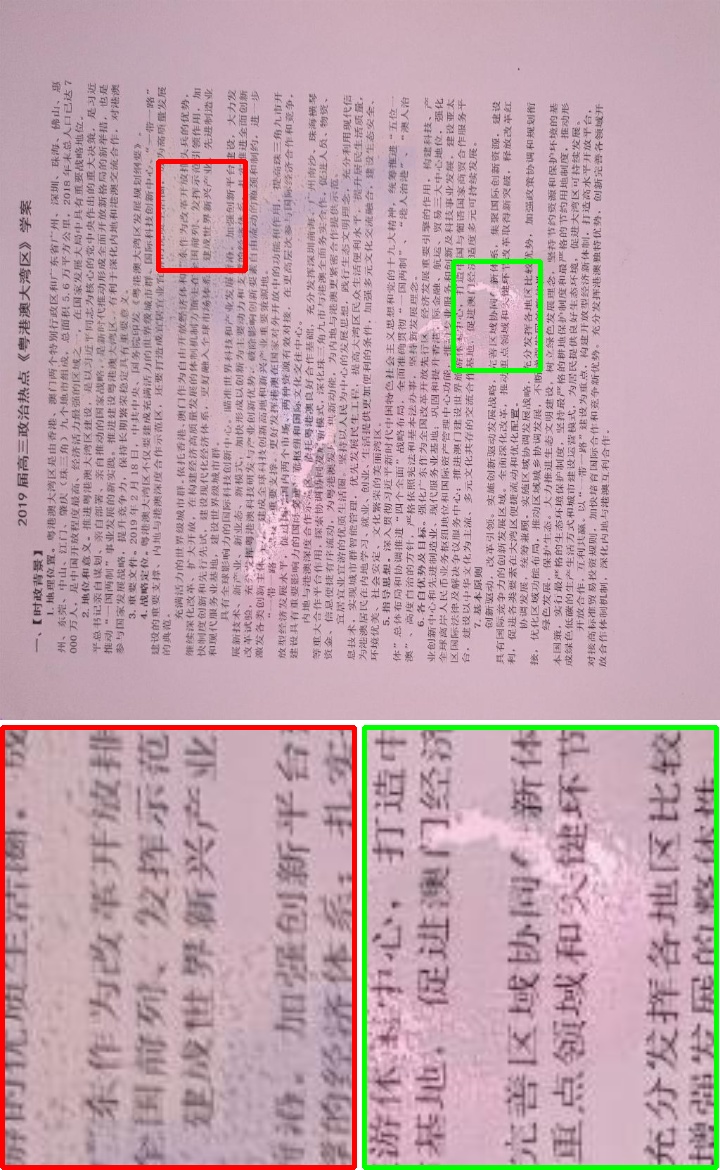}}
        \end{minipage}
        \hfill
        \begin{minipage}[b]{0.16\linewidth}
            \centering
            \centerline{\includegraphics[width=\linewidth]{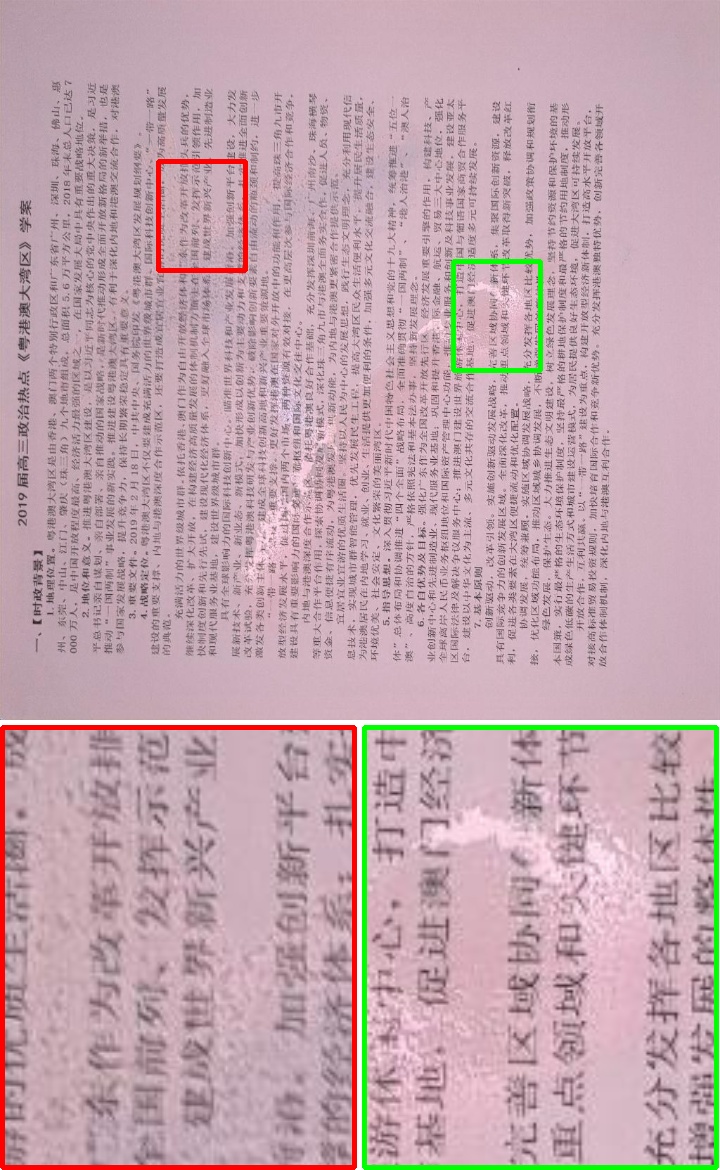}}
        \end{minipage}
        \hfill
        \begin{minipage}[b]{0.16\linewidth}
            \centering
            \centerline{\includegraphics[width=\linewidth]{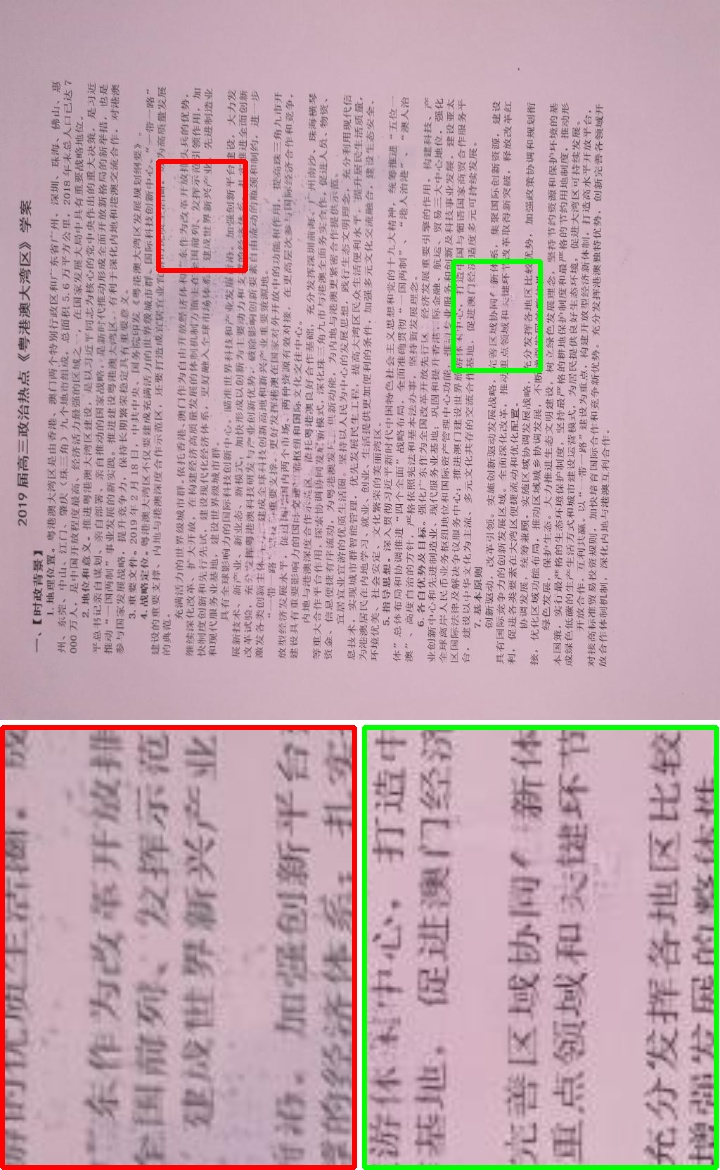}}
        \end{minipage}
        \hfill
        \begin{minipage}[b]{0.16\linewidth}
            \centering
            \centerline{\includegraphics[width=\linewidth]{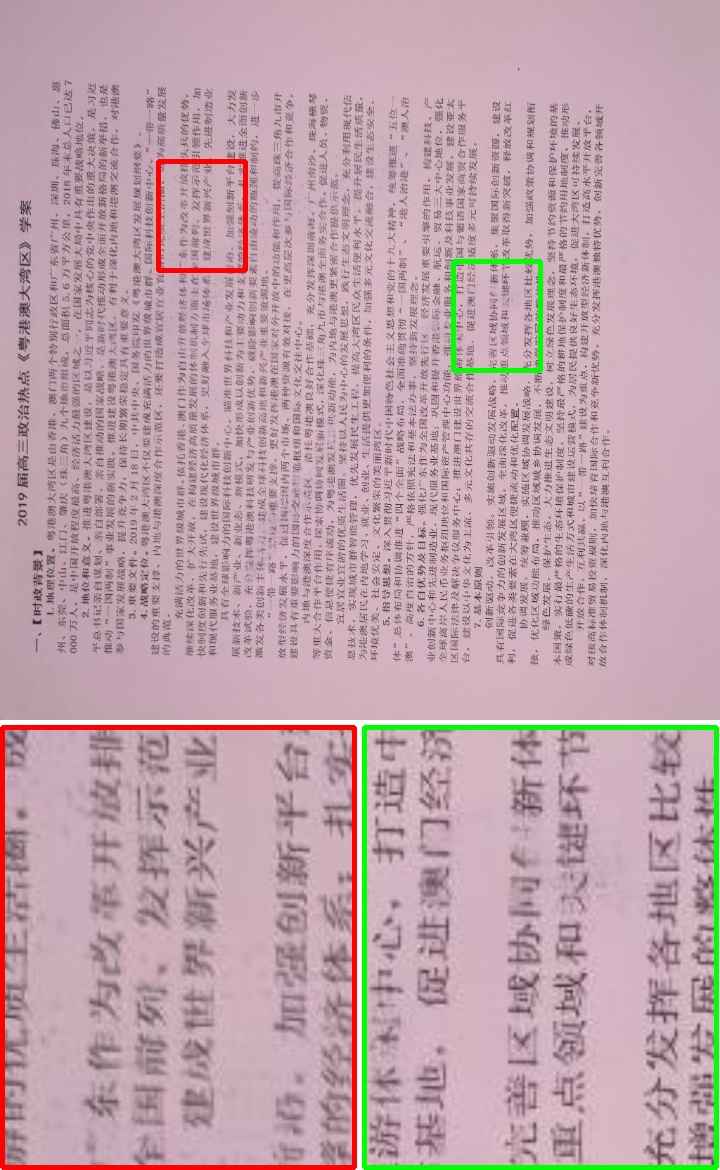}}
        \end{minipage}
        \hfill
        \begin{minipage}[b]{0.16\linewidth}
            \centering
            \centerline{\includegraphics[width=\linewidth]{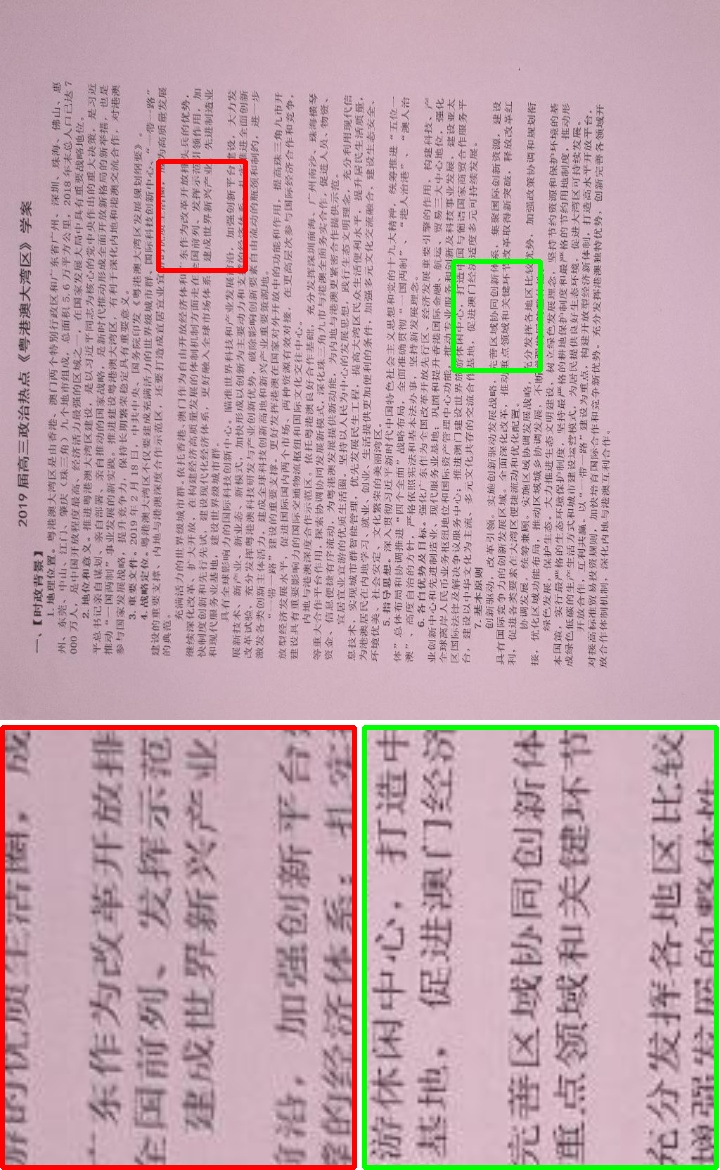}}
        \end{minipage}

    \end{minipage}
    \begin{minipage}[b]{1.0\linewidth}
        \begin{minipage}[b]{0.16\linewidth}
            \centering
            \centerline{\includegraphics[width=\linewidth]{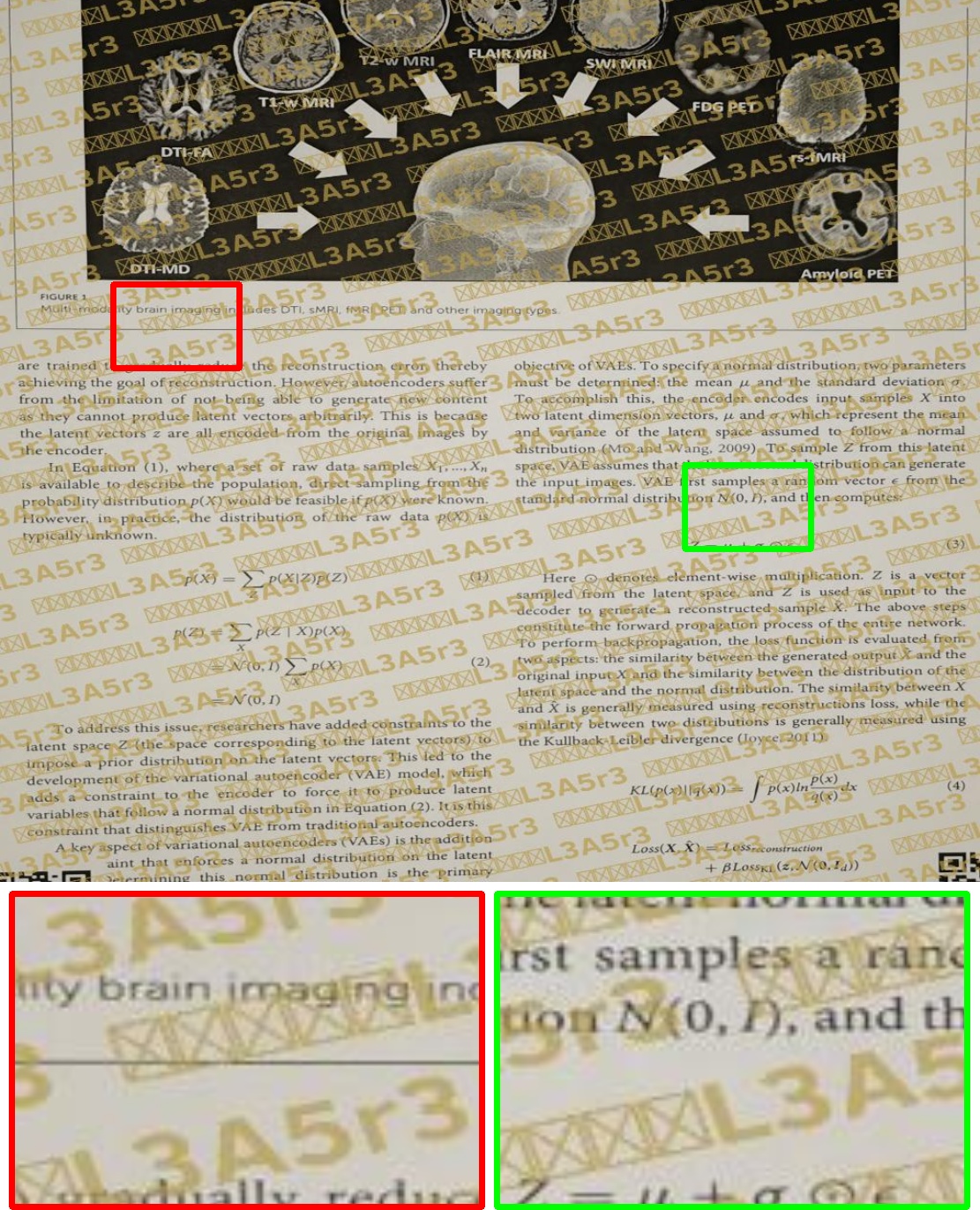}}
        \end{minipage}
        \hfill
        \begin{minipage}[b]{0.16\linewidth}
            \centering
            \centerline{\includegraphics[width=\linewidth]{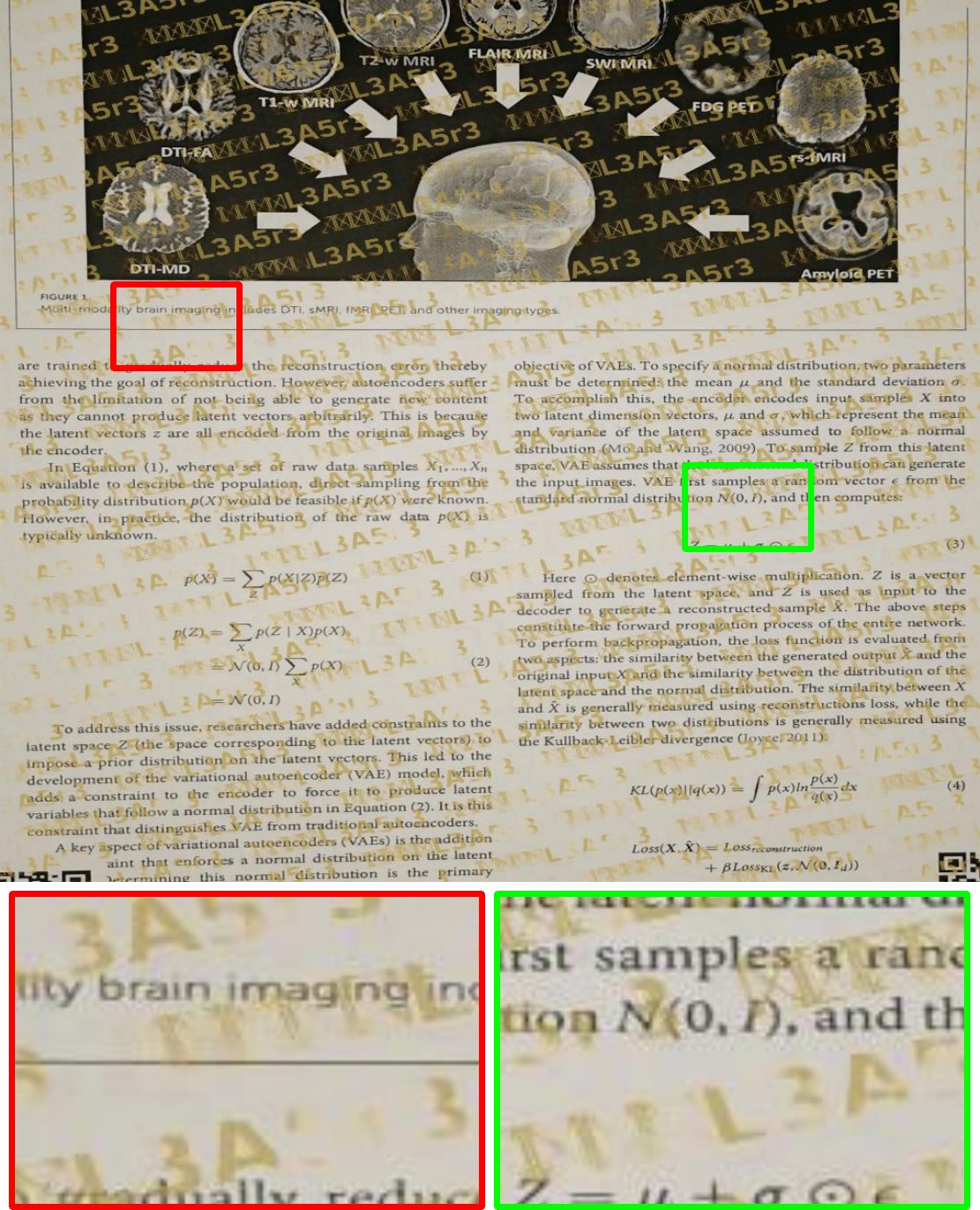}}
        \end{minipage}
        \hfill
        \begin{minipage}[b]{0.16\linewidth}
            \centering
            \centerline{\includegraphics[width=\linewidth]{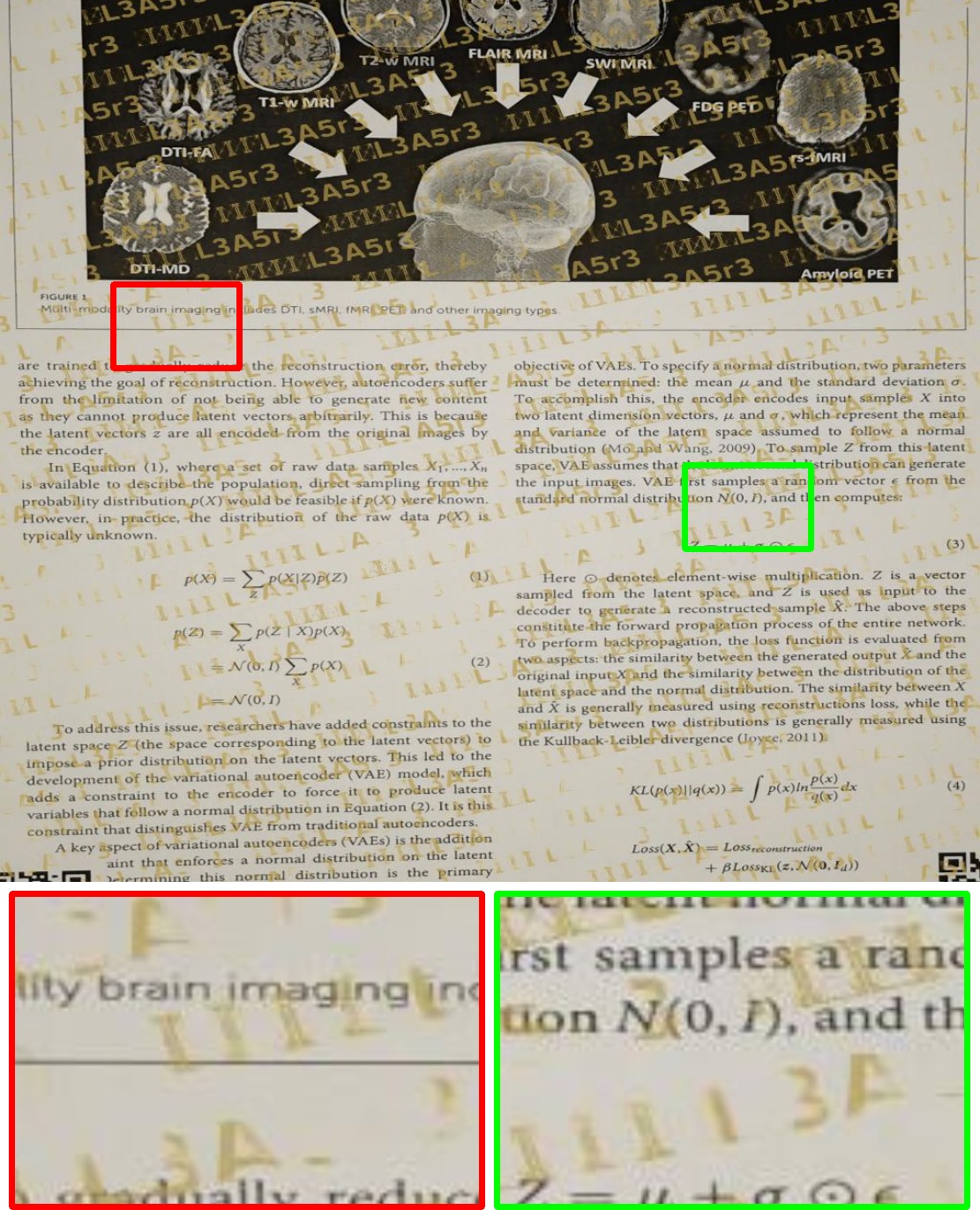}}
        \end{minipage}
        \hfill
        \begin{minipage}[b]{0.16\linewidth}
            \centering
            \centerline{\includegraphics[width=\linewidth]{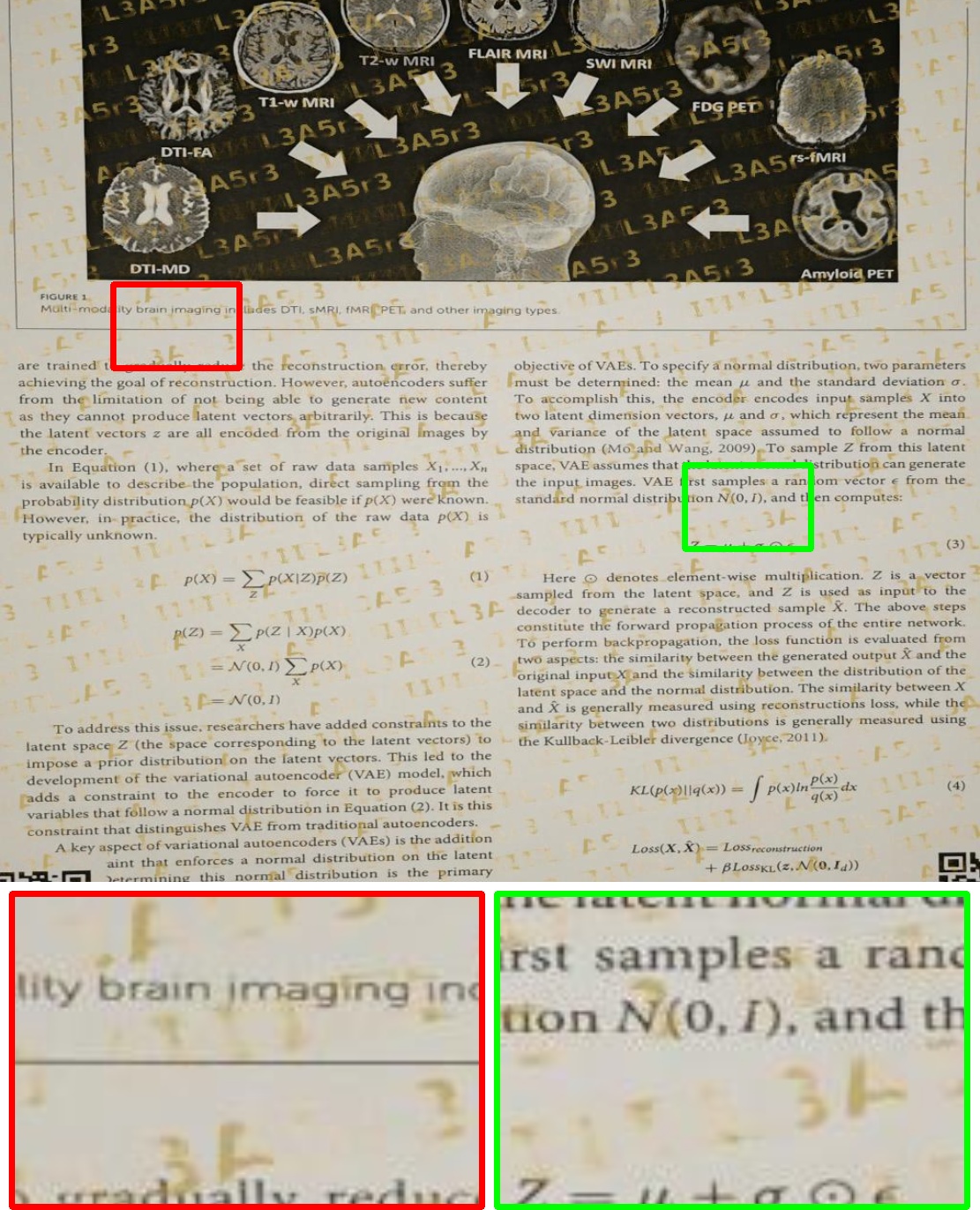}}
        \end{minipage}
        \hfill
        \begin{minipage}[b]{0.16\linewidth}
            \centering
            \centerline{\includegraphics[width=\linewidth]{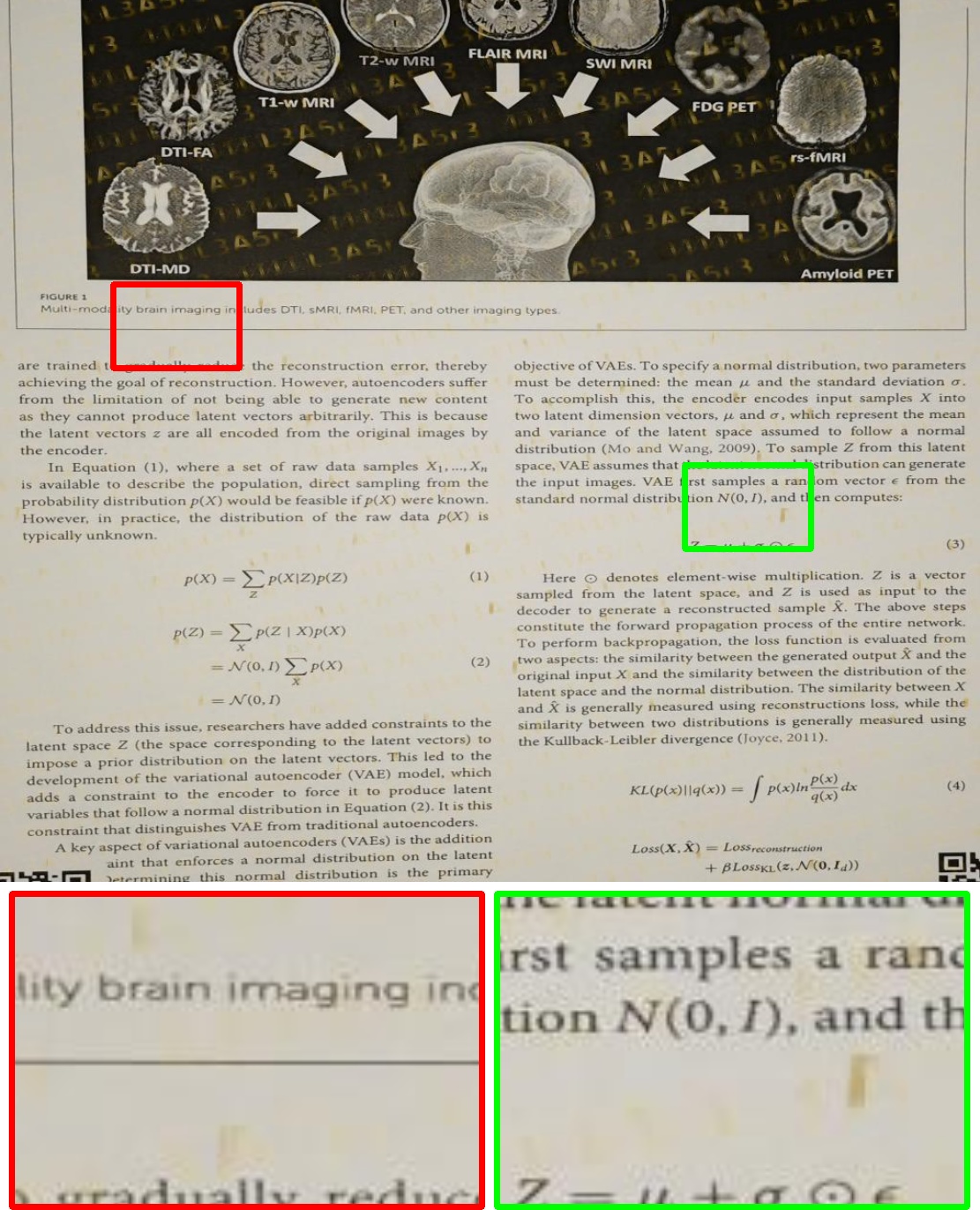}}
        \end{minipage}
        \hfill
        \begin{minipage}[b]{0.16\linewidth}
            \centering
            \centerline{\includegraphics[width=\linewidth]{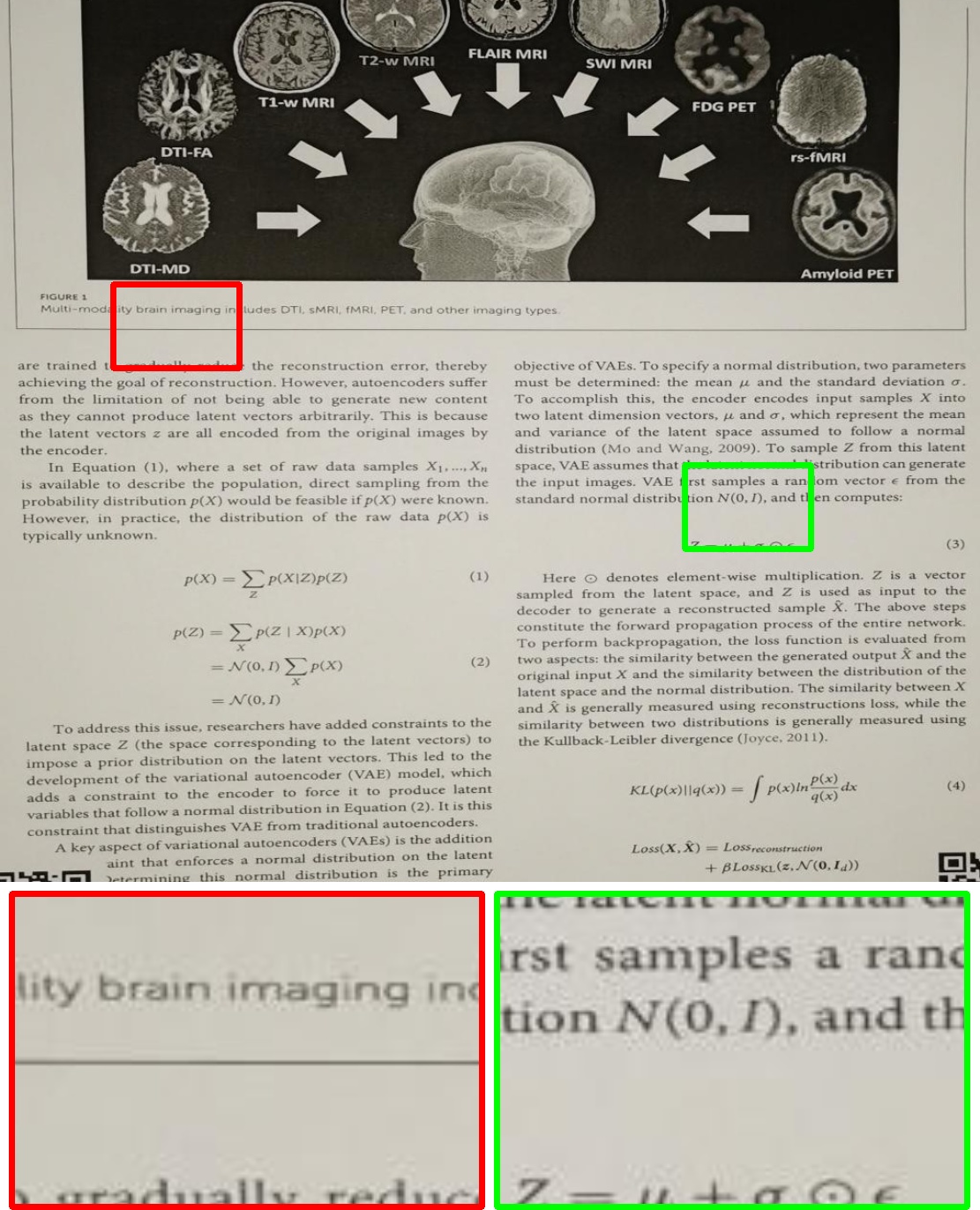}}
        \end{minipage}

    \end{minipage}

    \begin{minipage}[b]{1.0\linewidth}
        \begin{minipage}[b]{0.16\linewidth}
            \centering
            \centerline{\includegraphics[width=\linewidth]{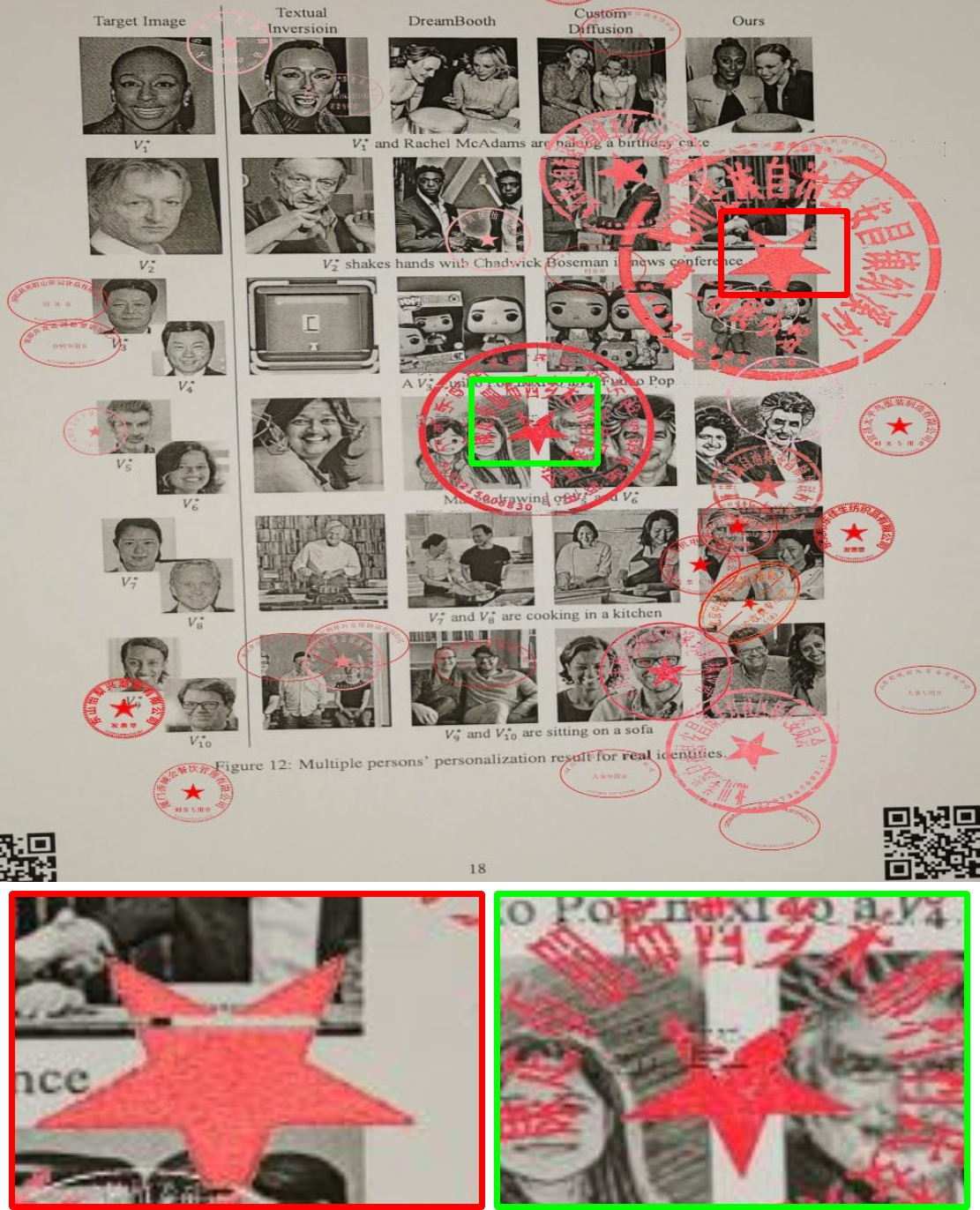}}
            \centerline{Input}
        \end{minipage}
        \hfill
        \begin{minipage}[b]{0.16\linewidth}
            \centering
            \centerline{\includegraphics[width=\linewidth]{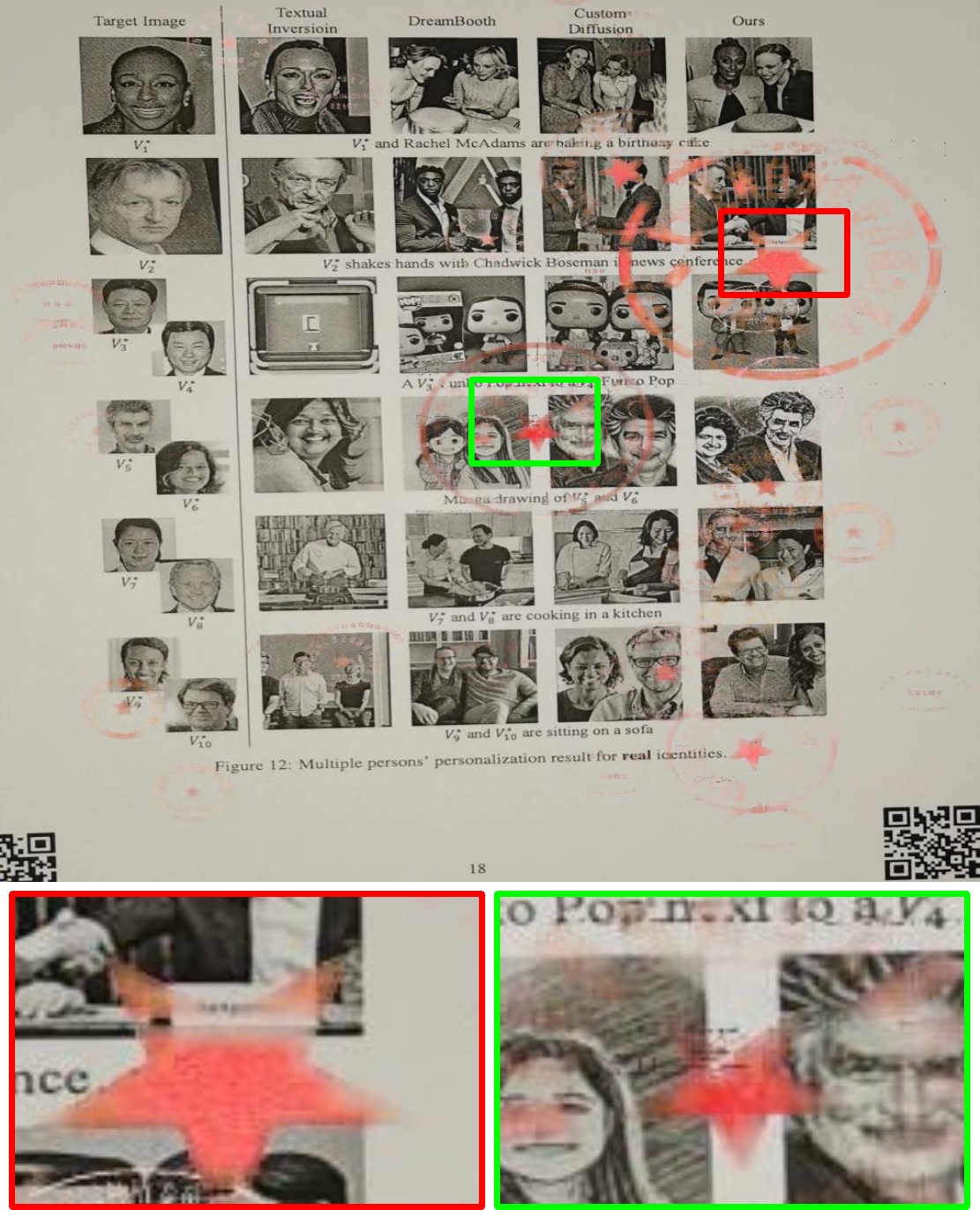}}
            \centerline{W/o both}
        \end{minipage}
        \hfill
        \begin{minipage}[b]{0.16\linewidth}
            \centering
            \centerline{\includegraphics[width=\linewidth]{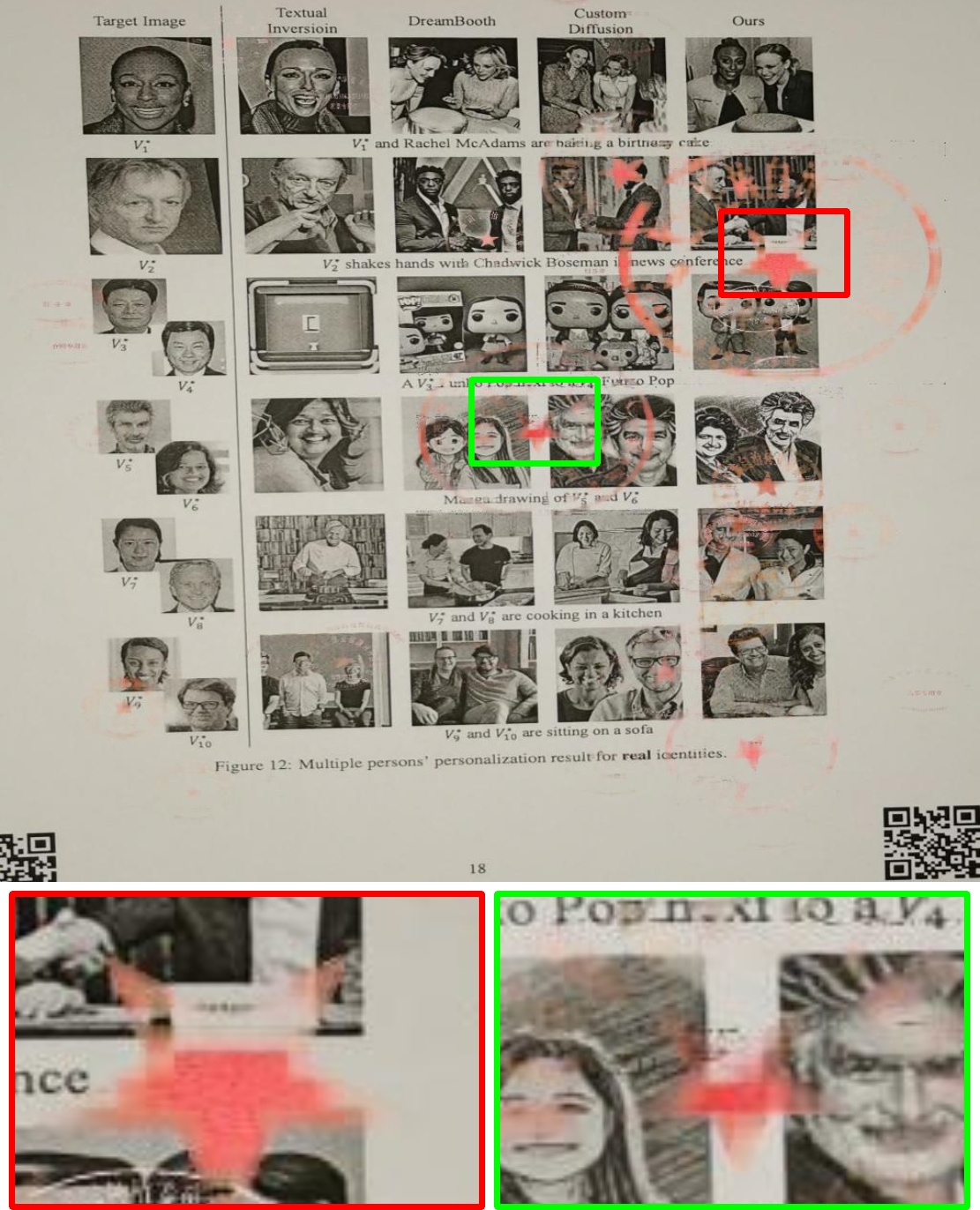}}
            \centerline{W/o SRTransformer}
        \end{minipage}
        \hfill
        \begin{minipage}[b]{0.16\linewidth}
            \centering
            \centerline{\includegraphics[width=\linewidth]{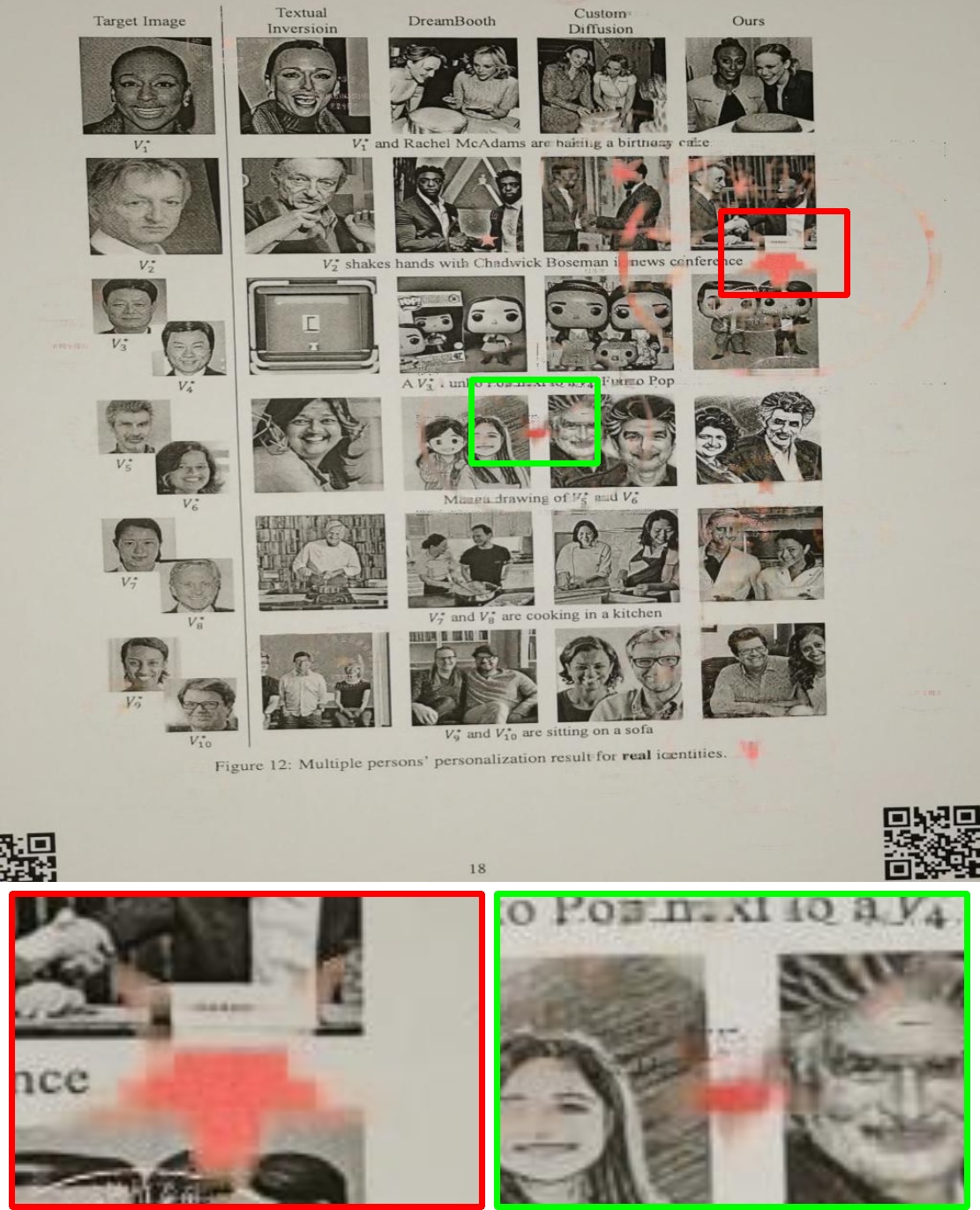}}
            \centerline{W/o DocMemory}
        \end{minipage}
        \hfill
        \begin{minipage}[b]{0.16\linewidth}
            \centering
            \centerline{\includegraphics[width=\linewidth]{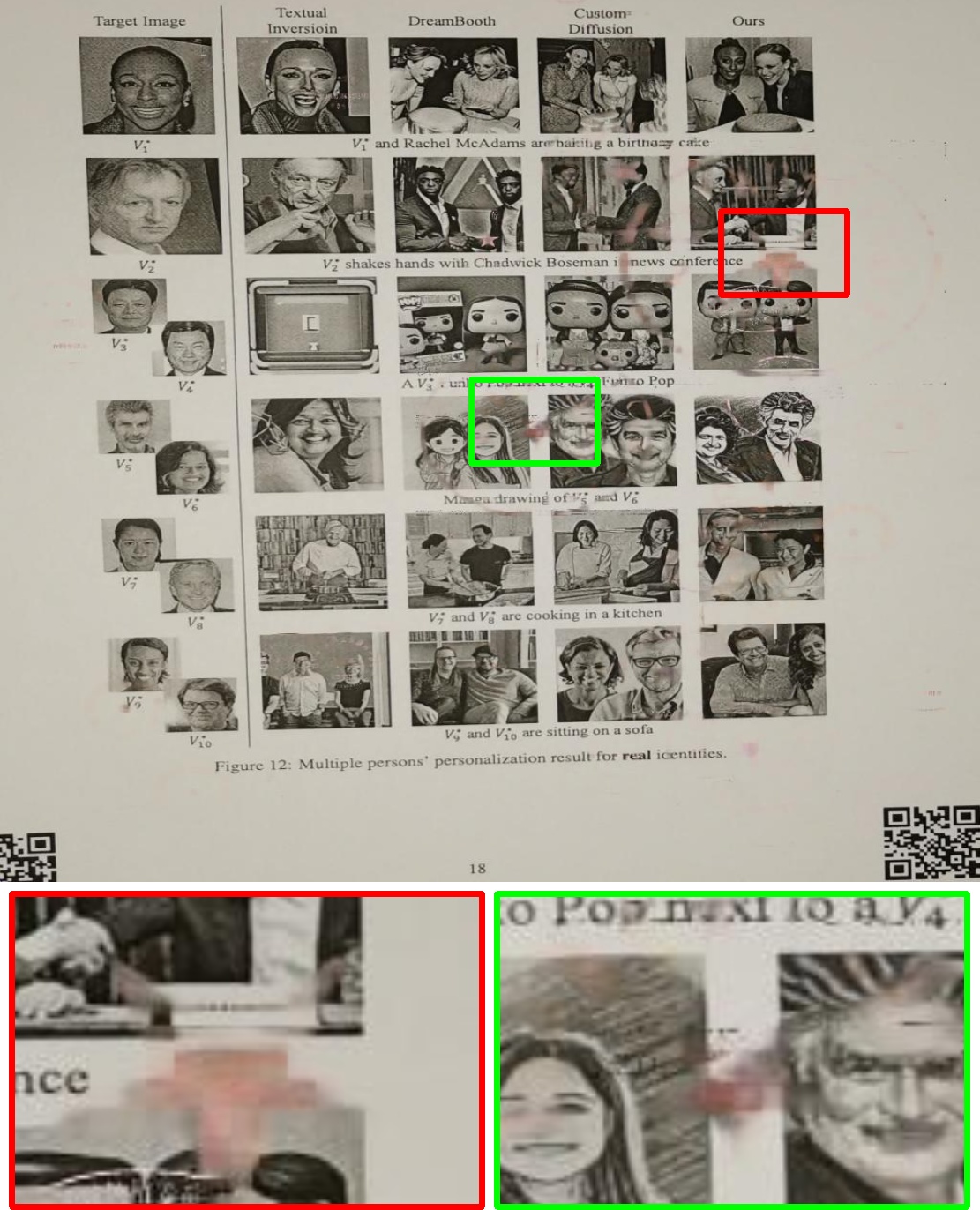}}
            \centerline{Full model}
        \end{minipage}
        \hfill
        \begin{minipage}[b]{0.16\linewidth}
            \centering
            \centerline{\includegraphics[width=\linewidth]{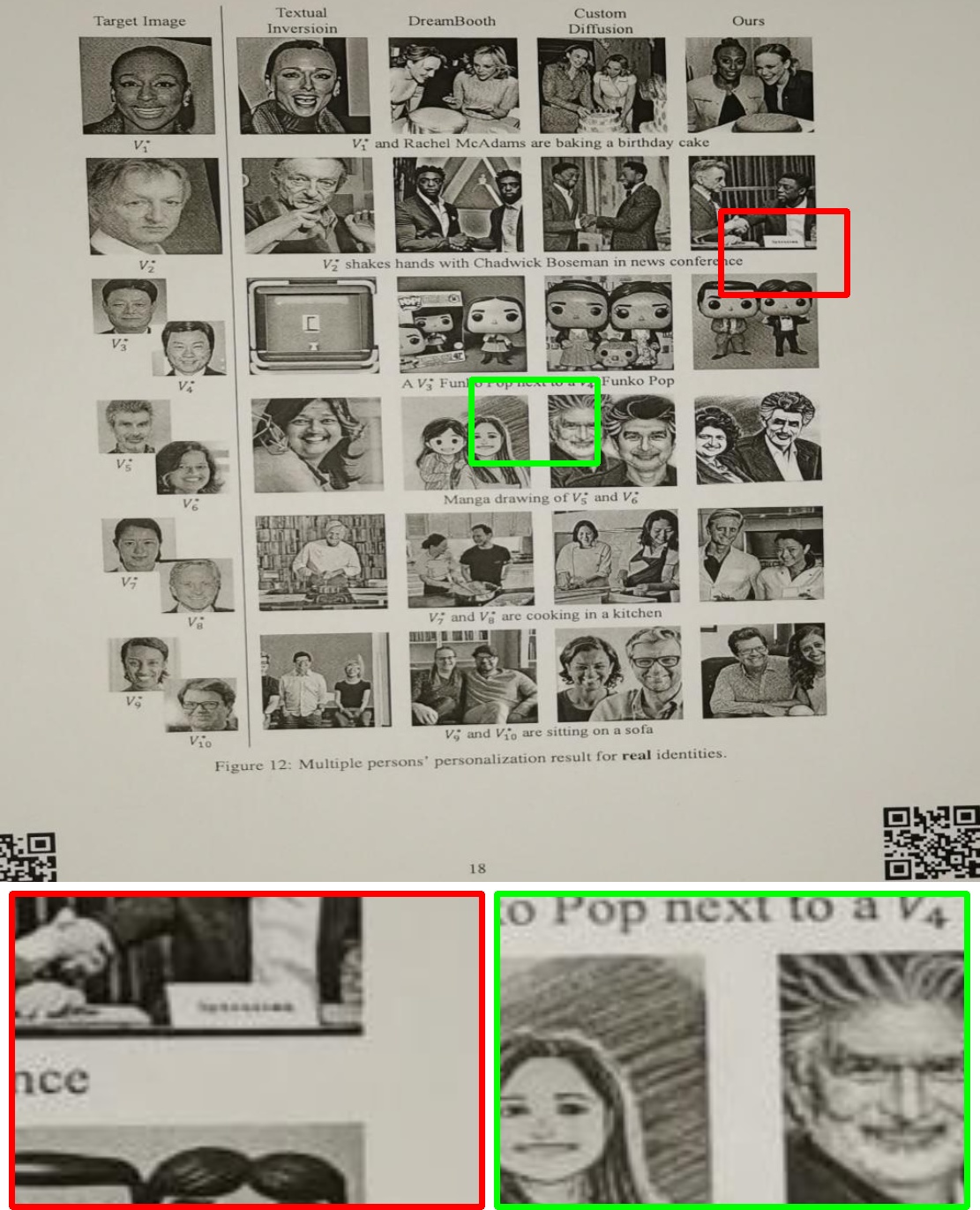}}
            \centerline{Target}
        \end{minipage}

    \end{minipage}

    \caption{
    Qualitative analysis of model predictions with different configurations on various datasets. 
    Each row illustrates the impact of ablating specific model components on the generated output. The first row corresponds to the StainDoc dataset, the second to StainDoc\_Mark, and the third to StainDoc\_Seal.
    }
    \label{fig:ablation}
    
\end{figure*}
StainRestorer is trained using a composite loss function that balances pixel-wise accuracy and the preservation of structural information. This is achieved by combining the Mean Squared Error (MSE) loss $\mathcal{L}_{MSE}$ and the Structural Similarity Index Measure (SSIM) loss $\mathcal{L}_{SSIM}$.

The $\mathcal{L}_{MSE}$ measures the average squared difference between the restored image ($\hat{I}$) and the target image ($I$):
\begin{equation}
\mathcal{L}_{MSE}(\hat{I}, I) = \frac{1}{N} \sum_{i=1}^{N} (\hat{I}(i) - I(i))^2,
\end{equation}
where $N$ represents the total number of pixels in the images. 

While $\mathcal{L}_{MSE}$ effectively captures pixel-level differences, it does not fully account for perceived visual quality, particularly concerning structural information. To address this, the $\mathcal{L}_{SSIM}$ is used, which compares images based on luminance, contrast, and structure. The $\mathcal{L}_{SSIM}$ is calculated locally over sliding windows and then averaged:
\begin{equation}
\mathcal{L}_{SSIM}(\hat{I}, I) = 1 - \frac{(2\mu_{\hat{I}}\mu_{I} + C_1)(2\sigma_{\hat{I}I} + C_2)}{(\mu_{\hat{I}}^2 + \mu_{I}^2 + C_1)(\sigma_{\hat{I}}^2 + \sigma_{I}^2 + C_2)},
\end{equation}
where $\mu_{\hat{I}}$ and $\mu_{I}$ are the mean pixel values of the restored image and the target image, respectively. $\sigma_{\hat{I}}$ and $\sigma_{I}$ are the standard deviations of the restored image and the target image, respectively. $\sigma_{\hat{I}I}$ is the covariance between the restored image and the target image. $C_1$ and $C_2$ are constants added to stabilize the division.

Finally, the total training loss function is a weighted sum of the MSE and SSIM losses:
\begin{equation}
\mathcal{L}_{total} = \mathcal{L}_{MSE}(\hat{I}, I) + \alpha \cdot \mathcal{L}_{SSIM}(\hat{I}, I),
\end{equation}
where $\alpha$ is a weighting factor controlling the relative importance of the SSIM loss. Empirically, $\alpha$ is set to 0.2. 
%%%%%--------------------------------------------------------------------------------------------------------------------------------%%%%%
\section{Experiment}
\label{sec:experiments}
\subsection{Datasets and Evaluation Metrics}
To assess the efficacy of document stain removal techniques, we utilize the StainDoc dataset. We partition the dataset, employing 4,502 image pairs for training and reserving the remaining 558 pairs for testing. 

To enhance the diversity of stain types within our evaluation, we introduce two supplementary datasets StainDoc\_Seal and StainDoc\_Mark for seals and marks removal. These datasets are generated leveraging the ground truth images from StainDoc as a foundation. Following the methodology outlined by Yang \etal~\cite{yang2023docdiff}, we generate these images with synthetic seals and printed font marks. This process uses 1,597 seal samples and 525 distinct font variations encompassing multilingual and alphanumeric characters. The training and testing splits are identical to those used for StainDoc.

StainDoc presents significant challenges due to its real-world complexities, featuring diverse stain patterns with irregular shapes, varying intensities, and non-uniform distributions. The overlap of stains with document content and the presence of multiple stain types further complicate the removal task. In contrast, StainDoc\_Seal and StainDoc\_Mark are comparatively easier to process, with controlled stain patterns, predictable shapes, and uniform stain types. Their synthetic nature results in less variability, making them more manageable for stain removal.

We evaluate the performance of our proposed StainRestorer model using 4 widely adopted image quality assessment metrics: PSNR (Peak Signal-to-Noise Ratio), SSIM (Structural Similarity Index Measure)~\cite{1284395}, MAE (Mean Absolute Error) and LPIPS (Learned Perceptual Image Patch Similarity)~\cite{8578166}.

\subsection{Implementation Details}
All experiments are conducted by the PyTorch framework and are executed on 2 NVIDIA V100 GPUs. During training, we employ a batch size of 4 and an initial learning rate of $2\times10^{-4}$. The AdamW optimizer~\cite{loshchilov2017decoupled}, with default parameters, is utilized for parameter updates. To facilitate dynamic learning rate adjustment, we incorporate a Cosine Annealing Learning Rate Scheduler~\cite{loshchilov2016sgdr}.

Training images are resized to a resolution of $256\times256$ pixels, while testing is conducted at a higher resolution of $1024\times1024$ pixels. To augment the training data and enhance model robustness, we apply random cropping, flipping, rotation, and mixup techniques.
\subsection{Comparisons with State-of-the-arts}
\begin{table}[ht]
\centering
\caption{Quantitative comparison of different methods on the StainDoc dataset, StainDoc\_Mark, and StainDoc\_Seal. The best results are highlighted in bold, and the second-best results are underlined.}
\begin{adjustbox}{width=\linewidth}
\begin{tabular}{c|cccc|cccc|cccc}
\toprule
\multirow{2}{*}{Method}&\multicolumn{4}{c|}{StainDoc}&\multicolumn{4}{c|}{StainDoc\_Mark}&\multicolumn{4}{c}{StainDoc\_Seal}
\\
\cmidrule(l){2-13}
& PSNR $\uparrow$ & SSIM $\uparrow$ & MAE $\downarrow$ & LPIPS $\downarrow$
& PSNR $\uparrow$ & SSIM $\uparrow$ & MAE $\downarrow$ & LPIPS $\downarrow$
& PSNR $\uparrow$ & SSIM $\uparrow$ & MAE $\downarrow$ & LPIPS $\downarrow$
\\
\midrule
Input         & 18.372          & 0.757          & 17.59          & 0.194          & 22.198          & 0.703          & 11.277         & 0.389          & 24.875          & 0.904          & 4.852          & 0.143          \\
DocTr++~\cite{feng2023deep}      & 14.012          & 0.43           & 32.057         & 0.692          & 14.927          & 0.348          & 29.097         & 0.767          & 16.861          & 0.518          & 18.634         & 0.534          \\
DocEnTR~\cite{souibgui2022docentr}& 15.173          & 0.514          & 34.955         & 0.84           & 16.095          & 0.545          & 29.643         & 0.775          & 17.424          & 0.558          & 20.126         & 0.686          \\
TextDIAE~\cite{souibgui2023text}      & 15.953          & 0.541          & 24.836         & 0.656          & 16.253          & 0.497          & 24.397         & 0.703          & 16.863          & 0.558          & 20.918         & 0.622          \\
Kligler~\cite{kligler2018document}       & 16.285          & 0.713          & 26.313         & 0.234          & 18.255          & 0.631          & 22.611         & 0.429          & 20.219          & 0.852          & 16.323         & 0.194          \\
DocDiff~\cite{yang2023docdiff}       & 16.662          & 0.607          & 25.609         & 0.359          & 17.646          & 0.658          & 23.842         & 0.435          & 12.891          & 0.458          & 44.894         & 0.673          \\
illtrtemplate~\cite{hertlein2023template}& 17.374          & 0.616          & 16.919         & 0.391          & 17.832          & 0.593          & 20.844         & 0.601          & 17.919          & 0.61           & 17.246         & 0.521          \\
GAN\_HTR~\cite{jemni2022enhance}      & 17.377          & 0.724          & 23.261         & 0.374          & 24.809          & 0.848          & 12.321         & 0.191          & 10.583          & 0.672          & 60.194         & 0.514          \\
DocTr~\cite{feng2021doctr}         & 17.87           & 0.597          & 20.812         & 0.601          & 16.866          & 0.613          & 16.736         & 0.38           & 16.879          & 0.613          & 16.734         & 0.383          \\
DeepOtsu~\cite{he2019deepotsu}      & 20.747          & 0.767          & 16.278         & 0.25           & 22.803          & 0.829          & 9.468          & 0.238          & 22.187          & 0.863          & 14.296         & 0.214          \\
DocProj~\cite{li2019document}       & 22.127          & 0.781          & 12.996         & 0.227          & 21.048          & 0.702          & 16.435         & 0.432          & 27.052          & 0.901          & 6.604          & 0.113          \\
DocNLC~\cite{wang2024docnlc}        & 22.45           & 0.781          & 11.378         & 0.232          & 22.031          & 0.72           & 13.866         & 0.407          & \underline{27.854}          & \underline{0.912}          & \underline{5.569}          & \underline{0.097}          \\
DocRes~\cite{zhang2024docres}        & 22.612          & 0.792          & 11.378         & 0.198          & 23.159          & 0.751          & 11.667         & 0.343          & 27.696          & 0.908          & 5.646          & 0.098          \\
GCDRNet~\cite{zhang2023appearance}       & 22.619          & 0.802          & 11.077         & 0.187          & \underline{26.194}          & \underline{0.893}          & \underline{6.629}          & \underline{0.121}          & 23.675          & 0.86           & 12.141         & 0.184          \\
DE-GAN~\cite{souibgui2020gan}         & 22.62           & \underline{0.819}          & 11.209         & \underline{0.154}          & 23.421          & 0.842          & 8.947          & 0.213          & 22.123          & 0.844          & 12.774         & 0.202          \\
UDoc-GAN~\cite{wang2022udoc}     & \underline{22.834}          & 0.803          & \underline{10.608}         & 0.189          & 25.94           & 0.88           & 6.888          & 0.142          & 26.391          & 0.881          & 8.399          & 0.142          \\
Ours     & \textbf{23.372} & \textbf{0.822} & \textbf{9.265} & \textbf{0.085} & \textbf{34.191} & \textbf{0.955} & \textbf{2.786} & \textbf{0.067} & \textbf{33.298} & \textbf{0.968} & \textbf{2.441} & \textbf{0.024}\\
\bottomrule
\end{tabular}
\end{adjustbox}
\label{tab:compare}
\end{table}
We compare the performance of the proposed StainRestorer against several state-of-the-art document enhancement and image restoration methods on the StainDoc, StainDoc\_Mark and Stain\_Seal datasets. 

\cref{tab:compare} presents the quantitative results of our comparison. StainRestorer consistently outperforms the other methods across all evaluation metrics. Notably, StainRestorer achieves state-of-the results in terms of PSNR, SSIM, and LPIPS, indicating its superior ability to restore image fidelity and perceptual quality. The MAE results further demonstrate StainRestorer's effectiveness in minimizing pixel-level errors during the stain removal process.

These results highlight the benefits of StainRestorer's hierarchical feature representation and memory-augmented architecture. By capturing stain characteristics at multiple levels of granularity and leveraging the spatial mapping capabilities of the SRTransformer, StainRestorer can effectively remove stains while preserving the underlying document content.

\cref{fig:compare} provides a qualitative comparison of the stain removal results. StainRestorer generates visually cleaner and more legible outputs compared to the other methods. It successfully removes various types of stains, including ink and beverage stains, while maintaining the integrity of the text and graphical elements. In contrast, the other methods either fail to completely eliminate the stains or introduce artifacts that degrade the visual quality of the restored documents.

Additional mean opinion scores and visual results are provided in the supplementary material.
\subsection{Ablation Studies}
\begin{table}[ht]
\centering
\caption{Ablation study on the StainDoc dataset, StainDoc\_Mark, and StainDoc\_Seal. The best results are highlighted in bold.}
\begin{adjustbox}{width=\linewidth}
\begin{tabular}{cc|cccc|cccc|cccc}
\toprule
\multicolumn{2}{c|}{Module} & \multicolumn{4}{c|}{StainDoc} & \multicolumn{4}{c|}{StainDoc\_Mark} & \multicolumn{4}{c}{StainDoc\_Seal} \\ 
\cmidrule(l){3-14} 
DocMemory           &SRTransformer            & PSNR $\uparrow$ & SSIM $\uparrow$ & MAE $\downarrow$ & LPIPS $\downarrow$  & PSNR $\uparrow$ & SSIM $\uparrow$ & MAE $\downarrow$ & LPIPS $\downarrow$   & PSNR $\uparrow$ & SSIM $\uparrow$ & MAE $\downarrow$ & LPIPS $\downarrow$   \\ \midrule
\ding{55} & \ding{55} &23.019 & 0.808     & 10.578    & 0.107     & 31.185    & 0.922     & 4.393     & 0.117     & 31.623    & 0.955     & 3.005     & 0.051     \\
\ding{51} & \ding{55} &23.106 & 0.799     & 9.766     & 0.123     & 31.493    & 0.914     & 4.323     & 0.121     & 32.303    & 0.961     & 2.72      & 0.04      \\
\ding{55} & \ding{51} &23.304 & 0.818     & 9.618     & 0.099     & 31.806    & 0.923     & 4.004     & 0.122     & 32.789    & 0.965     & 2.576     & 0.03      \\
\ding{51} & \ding{51} & \textbf{23.372} & \textbf{0.822} & \textbf{9.265} & \textbf{0.085} & \textbf{34.191} & \textbf{0.955} & \textbf{2.786} & \textbf{0.067} & \textbf{33.298} & \textbf{0.968} & \textbf{2.441} & \textbf{0.024}\\ \bottomrule
\end{tabular}%
\end{adjustbox}
\label{tab:ablation}
\end{table}
We conduct ablation studies to evaluate key components of StainRestorer. Specifically, we evaluate the effectiveness of the DocMemory module and its hierarchical structure, as well as the contribution of the SRTransformer.

\cref{tab:ablation} and~\cref{fig:ablation} present the results. Removing the DocMemory module leads to a notable performance drop across all metrics, highlighting its importance in capturing multi-level stain representations. Replacing the hierarchical structure of DocMemory with a single-level representation also results in degraded performance, emphasizing the benefit of progressively refining the understanding of stain characteristics.

The SRTransformer also proves crucial. Replacing it with another state-of-the-art Vision Transformer module~\cite{zamir2022restormer,zhang2024docres} significantly impacts the restoration quality, as evidenced by the lower PSNR, SSIM, and LPIPS scores. This demonstrates the effectiveness of SRTransformer's spatial mapping and attention mechanisms in precise stain removal while preserving document content.

These studies validate StainRestorer's design choices, emphasizing the synergistic contribution of the DocMemory module's hierarchical structure and the SRTransformer in achieving state-of-the-art performance for document stain removal.

\section{Conclusion}
\label{sec:conclusion}
This paper introduces StainRestorer, a deep learning model for high-fidelity document stain removal, and StainDoc, the first comprehensive dataset for this task comprising over 5,000 pairs of stained and clean document images with diverse stain types, severities, and backgrounds. StainRestorer leverages a novel hierarchical memory-augmented module (DocMemory) to capture multi-level stain representations and a Stain Removal Transformer (SRTransformer) to perform precise stain removal while preserving document content. Extensive experiments on StainDoc demonstrate StainRestorer's superior performance compared to existing document enhancement and image restoration methods. This work provides a valuable resource and foundation for future research in document stain removal, with potential applications in digital archiving and document analysis.

{\small
\bibliographystyle{ieee_fullname}
\bibliography{egbib}
}

\clearpage
\newpage
\appendix
\section{User Study}
While general objective image quality metrics like Naturalness Image Quality Evaluator (NIQE) offers valuable insights into the visual quality of images, there's a shortage of subjective metrics specifically tailored to assess the effectiveness of document image stain removal techniques. To address this gap and validate the effectiveness of our StainRestorer in practical applications, we conducted a user study focusing on perceptual quality improvements. We compared our enhanced images against those from state-of-the-art techniques applied to the same degraded originals, emphasizing how each method visually improved upon the degradations.

For comprehensive assessment, we selected five state-of-the-art methods detailed in the comparative experiments section of the main text. These methods represent a range of approaches to document image restoration and provide a robust benchmark for our proposed method. We then randomly selected 10 images from each of the three test sets (StainDoc, StainDoc\_mark, StainDoc\_seal), resulting in a total of 180 images for evaluation. These sets cover various degradation types, including stains, watermarks, and seals. The images were organized into groups, each consisting of one original input image with stain removal results, including our method and the five other methods. The methods were anonymized to prevent bias, and the order of the methods within each group was randomized.

Thirty participants (15 males and 15 females, aged 22-45) were recruited for the study. This group included both experts in image processing (10) and non-experts (20) to ensure a balanced evaluation. All participants reported normal or corrected-to-normal vision and were unaware of the experiment's purpose. Participants rated the six methods on a scale of 1 (worst) to 5 (best) for each image group based on the following criteria:
\begin{enumerate}
    \item \textbf{Stain Removal Effectiveness}:  Assessing how well each method removed stains from the document image, considering the completeness and uniformity of stain removal.
    \item \textbf{Naturalness of Background Restoration}: Evaluating how natural the restored background area appears compared to the surrounding regions, focusing on the absence of color inconsistencies, blurriness, or artifacts.
    \item \textbf{Texture Preservation}: Assessing the consistency of texture and fine details in the restored area compared to the original texture of the document.
\end{enumerate}

The scoring scale ranged from 1 (worst) to 5 (best), allowing participants to capture a spectrum of perceptible quality levels in the highlight removal results:
\begin{itemize}
\item 1 (Poor): Significant issues or distortions.
\item 2 (Fair): Visible flaws but some acceptable elements.
\item 3 (Average): Satisfactory overall, with most elements adequately processed.
\item 4 (Good): Well-processed, with only minor imperfections.
\item 5 (Excellent): Exceptional quality.
\end{itemize}

Participants assigned scores for each criterion independently, ensuring a thorough evaluation of the various aspects of document image restoration. We calculated the Mean Opinion Score (MOS) for each method by averaging the scores across all participants and images. \Cref{tab:userstudy} presents the final user study scores, demonstrating that our method consistently achieves the highest average score across all three test sets.
\begin{table}[ht]
\caption{Comparison of user study scores with other five state-of-the-art methods. The highest-scored results are highlighted in bold, while the second-best are underlined for emphasis.}
\centering
\adjustbox{width=\linewidth}{%
\begin{tabular}{lccc}
\toprule
Method                 & StainDoc          & StainDoc\_mark         & StainDoc\_seal         \\
\midrule
illtrtemplate~\cite{hertlein2023template}           & 1.20          & 1.13          & 1.17         \\
DocNLC~\cite{wang2024docnlc}         & 3.49          & 2.53     & \underline{3.77}          \\
GCDRNet~\cite{zhang2023appearance}  & 3.50   & \underline{3.68}         & 3.19          \\
DE-GAN~\cite{souibgui2020gan}    & \underline{3.53}          & 3.27          & 3.38          \\
UDoc-GAN~\cite{wang2022udoc}           & 3.20         & 3.59 &         3.45    \\
Ours                   & \textbf{4.33} & \textbf{4.85} & \textbf{4.81}  \\
\bottomrule
\end{tabular}
}
\label{tab:userstudy}
\end{table}

\end{document}